\documentclass[]{style}

\usepackage[toc,page,header]{appendix}

\usepackage{minitoc}
\usepackage{multirow}
\usepackage{graphicx}
\usepackage{array}
\usepackage{makecell}
\usepackage{framed}
\usepackage{hyperref}
\usepackage{amsmath} 
\usepackage[colorinlistoftodos]{todonotes}
\usepackage{longtable}
\usepackage{hhline}
\usepackage{fancyvrb}
\usepackage{fvextra}
\usepackage{CJKutf8}
\usepackage{multicol}
\usepackage{cleveref}
\usepackage{tablefootnote}
\usepackage{threeparttable}
\usepackage{tabularx}
\usepackage{mdframed}
\usepackage{subcaption}
\usepackage[usestackEOL]{stackengine}
\usepackage[numbers]{natbib}
\newcommand{\commentout}[1]{}
\renewcommand{\paragraph}[1]{\noindent\textbf{#1.}\hspace*{1em}}
\usepackage{enumitem}
\setlist[itemize]{leftmargin=15pt}

\usepackage{colortbl}
\PassOptionsToPackage{table}{xcolor}

\usepackage{marvosym}

\RequirePackage{xspace}
\makeatletter
\DeclareRobustCommand\onedot{\futurelet\@let@token\@onedot}
\def\@onedot{\ifx\@let@token.\else.\null\fi\xspace}

\def\eg{\emph{e.g}\onedot} 
\def\ie{\emph{i.e}\onedot}

\makeatother

\title{RoboBrain 2.5: Depth in Sight, Time in Mind.}

\author{BAAI RoboBrain Team}

\contribution{Please see \hyperref[sec:contributions]{Contributions and Author List} for more author details.}

\abstract{
We introduce \textbf{RoboBrain 2.5}, a next-generation embodied AI foundation model that advances general perception, spatial reasoning, and temporal modeling through extensive training on high-quality spatiotemporal supervision. Building upon its predecessor, RoboBrain 2.5 introduces two major capability upgrades. Specifically, it unlocks \textbf{Precise 3D Spatial Reasoning} by shifting from 2D pixel-relative grounding to depth-aware coordinate prediction and absolute metric constraint comprehension, generating complete 3D manipulation traces as ordered keypoint sequences under physical constraints. Complementing this spatial precision, the model establishes \textbf{Dense Temporal Value Estimation} that provides dense, step-aware progress prediction and execution state understanding across varying viewpoints, producing stable feedback signals for downstream learning. Together, these upgrades extend the framework toward more physically grounded and execution-aware embodied intelligence for complex, fine-grained manipulation. The code and checkpoints are available at project website: \url{https://superrobobrain.github.io}. \vspace{-3em}
}

\begin{document}
\maketitle

\vspace{-2.5em}
\begin{figure*}[h]
    \centering
    \includegraphics[width=0.97\linewidth]{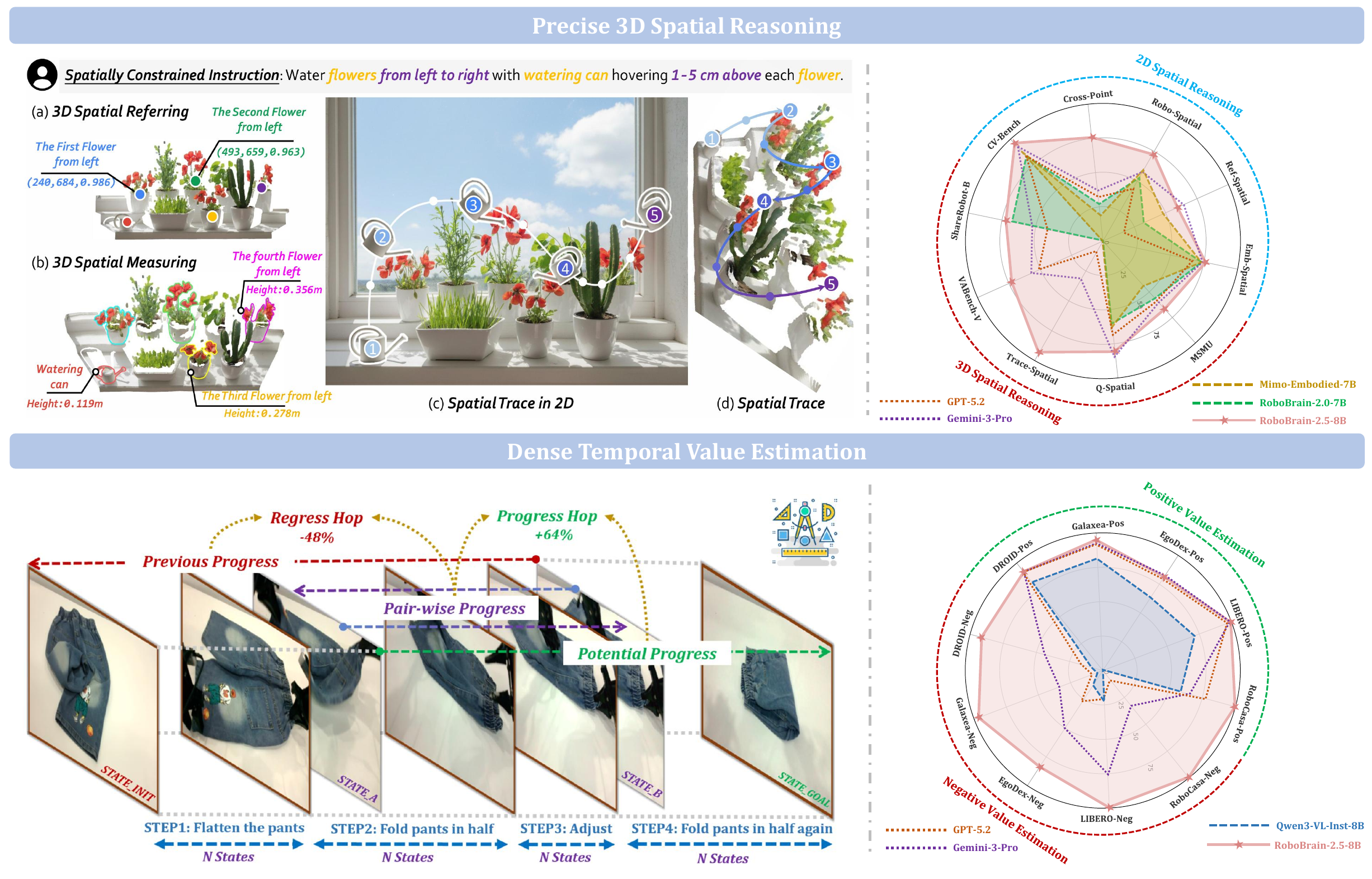}
    \caption{\textbf{New Features of RoboBrain 2.5.} Top: Precise 3D spatial reasoning with depth-aware grounding, metric measuring, and full manipulation trace generation under physical constraints. Bottom: Dense temporal value estimation for step-aware progress/regress prediction from state transitions across viewpoints and tasks; radar plots summarize performance gains on 2D/3D spatial and temporal benchmarks.}
    \label{fig:teaser}
\end{figure*}

\newpage
\tableofcontents
\newpage

\section{Introduction}

Embodied AI foundation models have rapidly advanced in bridging language, vision, and action, enabling the generation of actionable plans from natural language instructions and visual observations~\cite{cosmos-reason1, geminirobotics, pi05}. However, a critical gap persists. While these models often succeed in curated demonstrations, they frequently falter during rigorous real-world deployments. This reliability issue stems from the challenge of translating high-level semantic reasoning into {physically grounded manipulation}. Real-world tasks are unforgiving. They demand that robots respect absolute metric constraints, operate robustly under occlusions and viewpoint shifts, and continuously self-correct in a closed loop. Unfortunately, these precise physical capabilities remain beyond the reach of current semantic planners.
\vspace{0.3em}

These requirements expose two fundamental limitations in current generalist models. 
\textit{First, on the spatial dimension}, models suffer from ``metric blindness.'' Grounding is typically restricted to 2D pixel coordinates or weak topological representations~\cite{qwen25vl,spatialvlm,zhou2025roborefer}. Lacking absolute depth and scale information, such outputs inherently fail to ensure physical compliance. Specifically, they cannot guarantee millimeter-level clearance or generate collision-free 3D trajectories which are critical for precise interaction. 
\textit{Second, on the temporal dimension}, models usually operate as ``open-loop'' predictors. They treat action generation as a static sequence prediction task without an intrinsic mechanism to monitor execution progress. Relying on sparse external supervision such as success labels~\cite{ma2023eureka, ahn2024autort}, the agent remains oblivious to intermediate failures like slippage or regression. This limitation makes adaptive recovery impossible in long-horizon tasks.
\vspace{0.3em}

To bridge this gap, embodied foundation models must undergo a paradigm shift from semantic reasoners to physically-grounded agents. This evolution requires two precise upgrades. Spatial reasoning must advance from 2D pointing to precise 3D planning to satisfy metric constraints. Simultaneously, temporal modeling must shift from open-loop generation to dense value estimation to ensure closed-loop reliability.
\vspace{0.3em}

To realize this vision, we present \textbf{RoboBrain 2.5}. Building upon the robust general perception and reasoning capabilities of its predecessor~\cite{ji2025robobrain,team2025robobrain}, this next-generation model introduces critical upgrades to align internal representations with physical reality. Through large-scale training on high-quality spatiotemporal data, RoboBrain 2.5 achieves a comprehensive upgrade in core capabilities:
\vspace{0.3em}

\vspace{0.3em}
\begin{itemize}
    \item \textbf{Spatial: Depth in Sight (Precise 3D Spatial Reasoning).}  We extend the spatial interface from 2D grounding to depth-aware coordinate prediction and full manipulation trace generation. Instead of predicting a single target point, the model learns to output an ordered sequence of keypoints that describes the complete manipulation procedure, thereby naturally encoding spatial planning. This capability is built via a curriculum of three complementary skills: \textit{(1) 3D Spatial Referring} to localize objects; \textit{(2) 3D Spatial Measuring} to estimate absolute metric quantities (\eg, distance, clearance) required by physical constraints; and \textit{(3) 3D Spatial Trace Generation} to produce collision-free keypoint traces. Crucially, this is achieved by standardizing supervision into a decoupled $(u, v, d)$ representation convertible to 3D via camera intrinsics, leveraging large-scale, high-quality 3D supervision across diverse scenes.

    \vspace{0.3em}

    \item \textbf{Temporal: Time in Mind (Dense Temporal Value Estimation).}  In parallel, we establish a breakthrough in temporal modeling that provides immediate, step-aware feedback robust to viewpoint variations. The objective is to estimate the execution state (progress, stagnation, regression, or error) using only visual observations. We implement this by modeling general reward on multi-view expert trajectories using \textit{hop-normalized temporal transition labels}. This formulation normalizes progress by the remaining distance to the goal, producing bounded and stable supervision signals even with dense sampling. Furthermore, we employ multi-perspective fusion to aggregate value predictions, significantly improving robustness under occlusion. Consequently, RoboBrain 2.5 provides dense progress tracking that serves as a high-fidelity reward signal for downstream reinforcement learning.

    \vspace{0.3em}

    \item \textbf{Synergy and Impact.}
    Crucially, RoboBrain 2.5 integrates these physical capabilities without sacrificing the general interactive reasoning of the original architecture. By imparting ``Depth in Sight'' to ensure kinematic feasibility and ``Time in Mind'' to ensure execution robustness, our model successfully bridges the reliability gap. Extensive experiments on serious benchmarks demonstrate state-of-the-art performance. Furthermore, real-world evaluations confirm superior zero-shot robustness in contact-rich tasks, effectively translating demo-level success into deployment-level reliability.
    
\end{itemize}
\section{New Feature}
\label{sec:method}

Building upon the foundation of RoboBrain 2.0~\cite{team2025robobrain} and utilizing the Qwen3-VL architecture~\cite{qwen3vl}, RoboBrain 2.5 introduces two core enhancements that further advance physical intelligence.
Specifically, we first detail the concept of \textbf{Precise 3D Spatial Reasoning} (\Cref{subsec:3D_reasoning}), which encompasses three metric-grounded competencies,-spatial referring, measuring, and tracing—derived solely from monocular RGB inputs~\cite{zhou2025robotracer}.
We then describe \textbf{Dense Temporal Value Estimation}(\Cref{subsec:dense_temporal_value_estimation}), which learns a general-purpose, step-aware process modeling from multi-view RGB-only observations~\cite{tan2025robo}.

\subsection{Precise 3D Spatial Reasoning}
\label{subsec:3D_reasoning}
For embodied agents to interact effectively with the physical world, they must accurately interpret and act upon spatial information. This necessitates a deep understanding of object locations, inter-object relationships, and precise metric quantities from visual observations. To address these fundamental requirements, we introduce a robust framework for \textbf{Precise 3D Spatial Reasoning}.

\subsubsection{3D Spatial Referring, Measuring, and Tracing}

Embodied robots usually have to execute actions based on increasingly complex, spatially constrained instructions~\cite{zhou2025roborefer, tan2025roboosnext, tan2026action, abdolmaleki2025gemini, ji2025visualtrans,tan2025reason, bai2025embodied, bai2025towards}, such as ``\textit{Water flowers from left to right with watering can hovering 1–5 cm above each one}'' in \Cref{fig:teaser}, where recent data-scarce Vision-Language-Action (VLA) models fail to master.
In this case, it would be beneficial to generate a 3D positional sequence, named as \textbf{3D spatial trace}, as an intuitive bridge to interpret the instruction following procedure in 3D space and guide the generation of actual action trajectories for robots.
However, this surrogate task (\ie, \textbf{3D spatial tracing}) is inherently challenging as it requires multi-step, metric-grounded reasoning in complex 3D scenes.
To be specific, each reasoning step requires two key components:
\textbf{(1)} \textbf{3D spatial referring} to resolve spatial relationships and accurately localize objects involved in the trace generation (\eg, identifying flowers with their from left to right order and locating them).
\textbf{(2)} \textbf{3D spatial measuring} to understand absolute, real-world metric quantities related to the trace in captured scene (\eg, quantifying each flower's physical height and 1–5 cm height above each).
To this end, we equip RoboBrain 2.5 with these three key capabilities, enabling it to directly predict metric-grounded outputs from monocular images under spatial constraints for direct interaction with the 3D physical world.

\subsubsection{3D Task Formulation}

We formalize 3D spatial tracing as the process of predicting an ordered sequence of 3D points \(\tau = \{p_t\}_{t=1}^{T}\)—each point \(p_t = (u_t, v_t, d_t)\) comprising image-plane coordinates \((u_t, v_t)\) and absolute depth \(d_t\)—from visual inputs (\eg, RGB images) and textual instructions via vision-language models. 
The resulting trace \(\tau\) functions as a spatial plan for guiding entities (\eg, a robot end-effector or an object) to execute instructions. 
Crucially, these instructions typically encode both 3D spatial referring and 3D spatial measuring, often requiring multi-step compositional reasoning. 
For instance, in \Cref{fig:teaser}, the instruction ``\textit{Water flowers from left to right with watering can hovering 1-5 cm above each flower}'' necessitates determining the 3D positions and heights of all flowers in the scene. 
Although intermediate spatial cues (\eg, points identified through 3D spatial referring) may not coincide with the final keypoints used in the spatial trace, they provide essential evidence for multi-step reasoning—thereby enabling precise trace generation under spatial constraints at the start, the end, and along the trajectory.
At the core of our approach lies a task formulation designed to facilitate training and to leverage diverse data sources effectively. 
Rather than predicting 3D coordinates in the form \((x, y, z)\) in camera or world frame, we adopt a decoupled \((u, v, d)\) representation, which can be trivially projected to 3D coordinates using known camera intrinsics.
This formulation is especially advantageous in embodied scenarios where camera parameters are readily accessible, as it obviates the need for vision-language models to learn camera geometry implicitly. Such an approach streamlines training and enhances accuracy.
Furthermore, the \((u, v, d)\) representation can be straightforwardly projected into lower-dimensional subspaces. 
For instance, omitting \(d\) yields a 2D visual trace (\ie, a sequence of points in the image plane), while retaining only the start and end points produces 3D or 2D spatial referring data (if depth is further removed).
This flexibility not only promotes data reusability but also ensures compatibility with existing 2D datasets ~\cite{deitke2025molmo, zhou2025roborefer}, thereby boosting multi-task learning performance through co-training across complementary tasks and modalities.

\subsection{Dense Temporal Value Estimation}
\label{subsec:dense_temporal_value_estimation}

Effective execution of long-horizon manipulation tasks demands more than just a final success signal; it requires continuous, granular feedback to guide the agent through complex intermediate states~\cite{ma2023liv,alakuijala2024video,ma2024gvl,zhai2025vlac,chen2025sarm}. To address the limitations of sparse feedback, we introduce \textbf{Dense Temporal Value Estimation}, a vision-based mechanism that provides real-time, step-aware progress assessments as temporal value feedback, enabling robust closed-loop control and efficient RL.

\subsubsection{Hop-wise Progress Construction}
Central to our approach is the formulation of value estimation as task progress; thus, our model functions as a vision-language estimator designed to infer fine-grained, real-time progress from visual inputs. To guarantee generalizability across diverse embodiments and task families, we implement a three-stage data curation pipeline handling diverse data origins. This process spans from raw video segmentation to a systematic, hop-based labeling strategy, as detailed below:

\vspace{0.3em}

\textit{\textbf{Step-wise task progress discretization.}}
Given raw multi-view video trajectories, we first segment each expert trajectory into sub-tasks using human-annotated multi-view keyframes $\{K_0, K_1, \dots, K_N\}$, where $K_0$ is the initial observation, $K_N$ is the final success observation, and each $K_j$ is a set of synchronized multi-view keyframes. To obtain dense supervision, we perform adaptive sampling within each segment. For a trajectory with $L$ frames per view, we set a chunk size $C$ to determine the total number of sampled points and distribute them uniformly across the $N$ segments. The number of intermediate points $m$ within segment $[K_j, K_{j+1}]$ is:
\begin{equation}
\label{eq:sampling_frames}
m = \left\lfloor \frac{1}{N} \left\lfloor \frac{L}{C} \right\rfloor \right\rfloor.
\end{equation}
This yields a sequence of states $\mathcal{S} = \{s_0, s_1, \dots, s_M\}$, 
where each state $s_i$ is a set of synchronous multi-view visual observations.
We then define the ground-truth global progress as $\Phi(s_i) = i/M$.

\vspace{0.3em}

\textit{\textbf{Hop-based relative progress normalization.}}
A naive choice is to regress the progress gain $\Phi_{\delta}(s_p, s_q) = \Phi(s_q) - \Phi(s_p)$ between two states, but iterating such predictions accumulates error and can push the reconstructed $\Phi^{\star}(s)$ outside $[0,1]$. Instead, we introduce a hop-based formulation that learns \textit{relative-relative progress} and naturally supports dense temporal value estimation. Each training sample is a tuple $\mathcal{D}$ containing a task description $d_{\text{task}}$, the initial state $s_0$, the goal state $s_M$, a \textit{``BEFORE''} state $s_p$, an \textit{``AFTER''} state $s_q$, and a hop label $\mathcal{H}(s_p, s_q)$ that normalizes the progress from $s_p$ to $s_q$ relative to the full task span from $s_0$ to $s_M$. Given $\Phi(s_p)$ and $\Phi(s_q)$, we define:
\begin{equation}
\label{eq:hop_calculation}
\mathcal{H}(s_p, s_q) = 
\begin{cases} 
  \dfrac{\Phi(s_q) - \Phi(s_p)}{\Phi(s_M) - \Phi(s_p)} & \text{if } q \geq p \textsc{ (progress)} \\
  \\
  \dfrac{\Phi(s_q) - \Phi(s_p)}{\Phi(s_p) - \Phi(s_0)} & \text{if } q < p \textsc{ (regress)}.
\end{cases}
\end{equation}
This dynamically scales the supervision into $[-1, 1]$: for forward progress, the change is normalized by the remaining distance to the goal; for regression, by the distance already covered from the initial state. A key theoretical advantage is that, when global progress is reconstructed by iteratively applying predicted hops, the resulting $\Phi^{\star}(s)$ is guaranteed to remain strictly within $[0,1]$. Please refer to~\Cref{app:proof} for the proof.

\vspace{0.3em}

\textit{\textbf{Sampling strategy and data balancing.}}
For each trajectory, we construct a balanced set of hop-based training samples. Continuous hop values are first discretized into $N_{\text{hop}}$ hop bins. The temporal distance between the \textit{``BEFORE''} state $s_p$ and \textit{``AFTER''} state $s_q$ in each pair is then chosen from $N_{\text{dis}}$ distance bins within each hop bin, yielding in total $N_{\text{hop}} \times N_{\text{dis}}$ non-trivial transitions. To reduce bias toward static segments, we further introduce an additional fraction $\alpha$ of samples explicitly labeled as zero-hop (i.e., $\mathcal{H}(s_p, s_q) = 0$), constructed by selecting pairs $(s_p, s_q)$ whose progress change is below a small threshold $\epsilon$:
\begin{equation}
|\Phi(s_q) - \Phi(s_p)| \le \epsilon.
\end{equation}

\subsubsection{Multi-Perspective Progress Fusion}
\label{sec:mppf}
To mitigate error accumulation and ensure consistent accuracy, we fuse dense temporal value estimates from three complementary perspectives: incremental prediction, forward-anchored prediction, and backward-anchored prediction.

\vspace{0.3em}

\textit{Incremental Prediction} offers a fine-grained, step-by-step assessment. Refer to \Cref{eq:hop_calculation}, the predicted global progress $\Phi_{I}^{\star}(s_t)$ is recursively computed from the preceding state's progress $\Phi^{\star}(s_{t-1})$ and the predicted hop $\mathcal{H}^\star(s_{t-1}, s_t)$. Let $\Delta\Phi^{\star}_{t-1, t}$ be the estimated progress hop:
\begin{equation}
\label{eq:incremental_pred}
\Delta\Phi^{\star}_{t-1, t} = \begin{cases} 
  [1-\Phi^{\star}(s_{t-1})] \cdot \mathcal{H}^\star & \text{if } \mathcal{H}^\star \geq 0 \\
  \Phi^{\star}(s_{t-1}) \cdot \mathcal{H}^\star & \text{if } \mathcal{H}^\star < 0.
\end{cases}
\end{equation}
The incremental progress is then calculated as follow:
\begin{equation}
\label{eq:incremental_progress}
\Phi_{I}^{\star}(s_t) = \Phi^{\star}(s_{t-1}) + \Delta\Phi^{\star}_{t-1, t},
\end{equation}
where $\Phi_{I}^{\star}(s_t)$ is accumulated along the trajectory, initialized with $\Phi^{\star}(s_{0})=0$. While this method excels at capturing local dynamics, it is susceptible to the accumulation of prediction errors over long trajectories.
To counteract this drift, we introduce two global perspectives. \textit{Forward-Anchored Prediction} provides a stable global reference by anchoring to the initial state $s_\text{init}$, where progress is zero:
\begin{equation}
\label{eq:forward_pred}
\Phi_{F}^{\star}(s_t) = \mathcal{H^\star}(s_\text{init}, s_t).
\end{equation}
Conversely, \textit{Backward-Anchored Prediction} is anchored to the goal state $s_\text{goal}$, where progress is one. This approach offers high sensitivity near task completion:
\begin{equation}
\label{eq:backward_pred}
\Phi_{B}^{\star}(s_t) = 1 + \mathcal{H^\star}(s_\text{goal}, s_t).  
\end{equation}

\vspace{0.3em}

These three methods offer complementary strengths: local precision (incremental), initial stability (forward), and goal sensitivity (backward). We fuse them via averaging to obtain a robust final progress estimate:
\begin{equation}
\label{eq:fused_pred}
\Phi^{\star}(s_t) = \frac{1}{3}\left(\Phi_{I}^{\star}(s_t) + \Phi_{F}^{\star}(s_t) + \Phi_{B}^{\star}(s_t)\right).
\end{equation}
This fusion yields a more accurate and drift-resistant value signal. Please also refer to ~\cite{tan2025robo} for how to apply this kind of value signal for RL process.

\subsubsection{Bi-directional Consistency Checking}
\label{sec:robust-estimation}

While the multi-perspective fusion via averaging (\Cref{eq:fused_pred}) serves as a baseline, its naive application in online RL faces the risk of Out-of-Distribution (OOD) hallucination. Due to the inherent limitations of data coverage, it is impossible for the training set to encompass every corner of the state space. During RL, the policy inevitably explores unseen regions where dense temporal value estimation may yield spurious high signals, leading to ``reward hacking.''
To address these, we propose a bi-directional consistency checking strategy that leverages consistency as a proxy for reliability. This design is motivated by the observation that forward $\Phi^*_F$ and backward $\Phi^*_B$ predictions tend to diverge significantly under OOD observations, whereas they remain consistent in familiar states.

\vspace{0.3em}

\textit{\textbf{Consistency-Aware Weighting.}}
We first define the mean estimated progress $\bar{\Phi}^*(s_t) = (\Phi^*_F(s_t) + \Phi^*_B(s_t)) / 2$. To quantify uncertainty, we calculate a normalized discrepancy metric:
\begin{equation}
    \Delta_{\text{norm}}(s_t) = \frac{|\Phi^*_B(s_t) - \Phi^*_F(s_t)|}{\bar{\Phi}^*(s_t) + \epsilon},
\end{equation}
where $\epsilon$ is a small constant for numerical stability. Normalization by $\bar{\Phi}^*$ ensures that discrepancies are penalized more heavily during the early stages (where $\Phi$ is small), as precise guidance is critical initially. We then derive a confidence weight $w_t \in (0, 1]$ using a Gaussian kernel with sensitivity $\alpha$:
\begin{equation}
    w_t = \exp\left( -\alpha \cdot (\Delta_{\text{norm}}(s_t))^2 \right).
\end{equation}

\vspace{0.3em}

\textit{\textbf{Conservative State Update.}}
To prevent the policy from exploiting erroneous estimates in OOD scenarios, we employ a conservative update rule for the maintained progress state $\Phi^*(s_t)$ instead of \Cref{eq:fused_pred}:
\begin{equation}
    \Phi^*(s_t) = \Phi^*(s_{t-1}) + \frac{w_t}{2} \cdot \left( \bar{\Phi}^*(s_t) - \Phi^*(s_{t-1}) + \Delta\Phi^{\star}_{t-1, t} \right).
\end{equation}
This mechanism acts as a semantic filter: it ignores uncertain updates when $w_t \to 0$ (retaining $\Phi^*(s_{t-1})$) and fully trusts the estimate when consistency is high ($w_t \to 1$).

\section{Training Data}
\begin{figure*}[h!]
  \centering
  \setlength{\abovecaptionskip}{0.5em}
  \setlength{\belowcaptionskip}{0em}
  \includegraphics[width=1.0\linewidth]{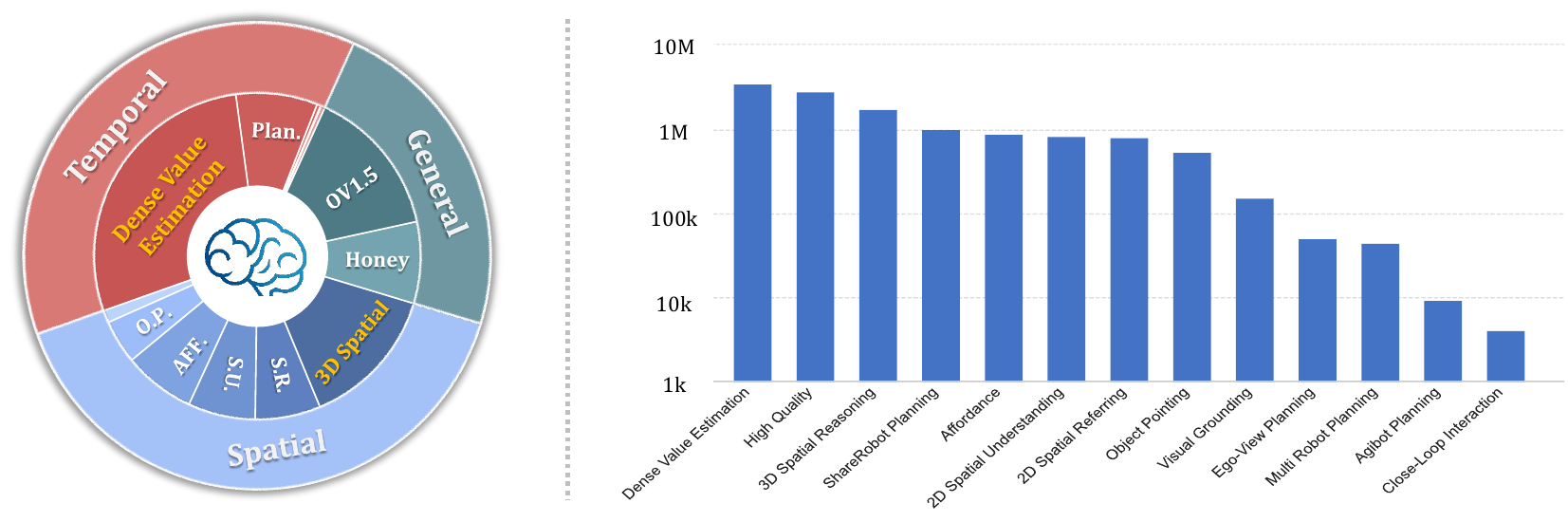}
  \caption{\textbf{Training Data Distribution for RoboBrain 2.5.} The left pie chart illustrates the hierarchical composition of the dataset, structured into Temporal (red), General (teal), and Spatial (blue) domains. The right bar chart displays the sample count for each specific sub-task on a logarithmic scale, highlighting the extensive scale of Dense Value Estimation, High-Quality General Data, and 3D Spatial Reasoning.}
  \label{fig:train_data}
\end{figure*}

As shown in \Cref{fig:train_data}, RoboBrain 2.5 is trained on a diverse and extensive dataset designed to enhance its capabilities in spatial understanding, temporal modeling and causal reasoning in embodied settings. Specifically, we construct a unified corpus of approximately 12.4M high-quality samples, categorized into three core domains: (1) \textit{General MLLM Data} for robust semantic perception; (2) \textit{Spatial Reasoning Data} spanning 2D perception to metric-aware 3D tracing; and (3) \textit{Temporal Prediction Data} for hierarchical planning and dense value estimation. This mixture strategically balances large-scale web knowledge with fine-grained physical world interactions to bridge the gap between high-level reasoning and low-level control.

\label{sec:train_data}

\subsection{General MLLM Data}
\label{sec:general_data}

\textbf{High-Quality General Data.} To establish a robust foundation for general visual perception and reasoning, the general training dataset for RoboBrain 2.5 incorporates approximately 2.83 million high-quality samples. These are primarily sourced and refined from two state-of-the-art open-source collections: Honey-Data-1M~\cite{zhang2025beehighqualitycorpusfullstack} and LLaVA-Onevision-1.5-Instruct-Data~\cite{an2025llava}.
\textbf{(1) Honey-Data-1M Processing.} We utilize Honey-Data-1M~\cite{zhang2025beehighqualitycorpusfullstack} as a key data source, which provides a diverse set of visual-language instructions designed to enhance multimodal understanding. To align the response style with our embodied agent's requirements for concise and direct execution commands, we truncated the extensive Chain-of-Thought (CoT) reasoning components, retaining only the final answers to streamline the supervision signal for direct instruction following.
\textbf{(2) LLaVA-Onevision Data Refinement.} We further integrate LLaVA-Onevision-1.5-Instruct-Data~\cite{an2025llava}, a comprehensive dataset covering a wide array of visual tasks including OCR, math, and general VQA. To strictly focus on vision-centric capabilities, we first filtered out all text-only samples. To address data imbalance, we applied balanced sampling across each visual-based subclass. Furthermore, to optimize training efficiency and context window utilization, we employed a sample packing strategy where shorter training samples are concatenated. This results in a more uniform sequence length distribution, primarily falling within the 2048 to 8192 token range.
\textbf{(3) De-duplication and Merging.} Given the overlap in data sources between these two repositories, we conducted a rigorous de-duplication process to prevent redundancy and data leakage. We filtered the combined pool based on both image similarity and question-answer textual similarity. The final curated dataset consists of 2.83M unique, high-quality multimodal instruction-following samples.

\subsection{Spatial Reasoning Data}

\textbf{Visual Grounding.}
The visual grounding dataset is constructed to enhance multimodal understanding through precise object-level localization, leveraging the extensive annotations from LVIS~\cite{gupta2019lvis}. We carefully curate 152K high-resolution images from LVIS, ensuring broad coverage of diverse object categories and complex visual scenes. Each object annotation is converted into standardized bounding box coordinates \((x_1,y_1,x_2,y_2)\) representing the top-left and bottom-right corners, enabling consistent spatial referencing. To facilitate rich visual dialogue, we generated 86K conversational sequences, each containing multiple rounds of QA pairs that progressively explore visual relationships, attribute reasoning, and contextual understanding. The dataset maintains a balanced distribution across object categories while preserving challenging cases of occlusion, viewpoint variation, and rare instances to support robust visual grounding. 

\vspace{0.3em}
\textbf{Object Pointing.}
The object pointing dataset is constructed to enable RoboBrain 2.5 to identify the locations of specified objects through pointing within an image. We leverage the Pixmo-Points~\cite{deitke2024molmo} dataset, which includes 2.3M point annotations across 223K images as our data source. However, direct utilization of Pixmo-Points data for RoboBrain 2.5 training presents challenges due to densely repeated object instances (\eg, books on a shelf). To address this, we implement a two-step filtering process: (1) we discard annotations with more than ten labeled points to simplify training, and (2) we use GPT-4o~\cite{gpt4o} as a scene analyzer to select only indoor-relevant objects, such as kitchenware, furniture, and decorations, excluding irrelevant or outdoor scenes. This process yields 190K QA pairs for 64K images with reduced clutter, making the data more suitable for embodied contexts. 
To construct QA pairs for pointing tasks, we construct 28 human-designed templates, such as ``\textit{Point out all instances of \{label\} in the image.}'' or ``\textit{Help me find \{label\} in the image by pointing to them.}'' Here, \textit{\{label\}} refers to object categories from the annotations. Templates are randomly selected to ensure linguistic diversity and improve the model's generalization ability in referencing tasks.
For object reference pointing, we incorporate object reference data sourced from RoboPoint~\cite{yuan2024robopointvisionlanguagemodelspatial}, which includes 347K QA annotations across 288K images. To address the potential issue of excessive points hindering training convergence, we randomly sample up to ten points per question. Additionally, all coordinates are converted into the normalized values to better support RoboBrain 2.5 training.

\vspace{0.3em}
\textbf{Affordance.}
The affordance dataset focuses on understanding object functionality and spatial vacant areas for placement. For object affordance recognition, we utilize part-level annotations from PACO-LVIS~\cite{ramanathan2023paco}, covering 75 object categories and 200 part categories across 46K images. Bounding boxes and segmentation masks are extracted for both whole objects and their functional parts. These annotations are transformed into bounding box coordinates \((x_1,y_1,x_2,y_2)\), serving as ground truth labels for affordance prediction tasks. Questions are constructed using GPT-4o~\cite{gpt4o} to query object functionality and part usage, \eg, ``\textit{Which part of a handbag can be grasped to carry it?}'' for the handle of a handbag. For whole-object affordances, questions avoid naming the object directly, such as ``\textit{What device can be moved to control the cursor on a screen?}'' for a mouse (computer equipment). This automatic process results in 561K QA pairs.
For spatial affordance learning, we include region reference data from RoboPoint~\cite{yuan2024robopointvisionlanguagemodelspatial}. This dataset consists of 270K images with 320K QA pairs and 14 spatial relationship labels. Each annotation is converted into a set of the normalized coordinates \([(x_1,y_1),(x_2,y_2),...]\), and ground truth points are resampled to a maximum of ten points per answer for optimization. This dataset enables RoboBrain 2.5 to reason about spatial affordances for object placement in real-world settings.

\vspace{0.3em}
\textbf{Spatial Understanding.}
To enhance RoboBrain 2.5’s spatial reasoning, we present the Spatial Understanding Dataset, comprising 826K samples. This dataset emphasizes object-centric spatial attributes (\eg, position, orientation) and inter-object relations (\eg, distance, direction), covering both qualitative and quantitative aspects. 
It covers 31 distinct spatial concepts, substantially surpassing the $\sim$15 typically found in previous datasets.
We partially adopt the RefSpatial~\cite{zhou2025roborefer} pipeline to construct 2D web image and 3D video datasets via automated template- and LLM-based generation: 
\textbf{(1) 2D web images} aim to provide core spatial concepts and depth perception across diverse indoor and outdoor scenes. To bridge scale and category gaps between these domains, we utilize the large-scale OpenImage~\cite{kuznetsova2020open} dataset. Since direct 3D reasoning from 2D images is challenging, we convert them into pseudo-3D scene graphs. Specifically, after filtering 1.7M images to 466K, we first use RAM~\cite{zhang2024recognize} for object category prediction and GroundingDINO~\cite{liu2024grounding} for 2D boxes Detection. Then we enhance using Qwen2.5-VL~\cite{qwen2.5-vl} and a heuristic method to generate hierarchical captions given the 2D bounding box, ranging from coarse (\eg, ``cup'') to fine-grained (\eg, ``the third cup from the left''). This enables unambiguous spatial referring in cluttered environments and captures both coarse and fine-grained spatial references. Next, we use UniDepth V2~\cite{piccinelli2025unidepthv2} and WildeCamera~\cite{zhu2023tame} for depth and camera intrinsics to enable 3D point cloud reconstruction. Finally, combining this with object boxes from GroundingDINO~\cite{liu2024grounding} and masks from SAM 2.1~\cite{ravi2024sam}, each scene graph includes object labels, 2D boxes, instance masks, and object-level point clouds, yielding axis-aligned 3D boxes. Object captions serve as nodes, and spatial relations form the edges. QA pairs are generated via templates and LLMs (\eg, QwQ~\cite{qwq32b}), including object-location questions derived from the hierarchical captions.
\textbf{(2) scanning datasets} integrates multimodal 3D scene understanding data from five original datasets: MMScan~\cite{mmscan}, 3RScan~\cite{3rscan}, ScanQA~\cite{scanqa}, SQA3D~\cite{sqa3d}, and SpaceR~\cite{ouyang2025spatial}. We conduct template-based question filtering through rigorous data processing to ensure task relevance, perform multi-stage quality screening (\eg, consistency checks, outlier removal), and standardize all formats into a unified representation. This curation enables fine-grained environmental perception with enhanced reliability, supporting tasks ranging from object localization to complex spatial reasoning in 3D scenes.
\textbf{(3) 3D embodied videos} focus on fine-grained spatial understanding in indoor environments. We leverage the CA-1M~\cite{lazarow2024cubify} dataset, filtering 2M frames to 100K high-quality ones. Compared to 2D, the availability of accurate 3D bounding boxes allows us to construct richer scene graphs with more diverse spatial relations, thereby generating more quantitative QA pairs (\eg, size, distances). 

\vspace{0.3em}
\textbf{Spatial Referring.}
After enhancing foundational 3D spatial understanding, we extend these capabilities to physical-world interactions by introducing the Spatial Referring Dataset~\cite{zhou2025roborefer}, consisting of 802K samples. Unlike prior datasets in visual grounding or object pointing, which often deal with ambiguous or multiple referents, this dataset targets a single unambiguous target, aligning with robotic applications such as precise pick-and-place that demand accurate object identification and localization.
Following the RefSpatial~\cite{zhou2025roborefer} construction pipeline, for location data, we sample caption-point pairs from scene graphs built on 2D web images (OpenImage~\cite{kuznetsova2020open}) and 3D embodied videos (CA-1M~\cite{lazarow2024cubify}), using hierarchical captions. For placement data, we leverage fully annotated 3D datasets to generate top-down occupancy maps encoding object positions, orientations, and metric spatial relations (\eg, ``10cm right of the chair''), facilitating accurate spatial referring.

\vspace{0.3em}
\textbf{3D Spatial Reasoning (RoboBrain 2.5 New Feature).} 
To equip the model with robust 3D spatial reasoning capabilities for tasks such as 3D spatial referring, measuring, and tracing, we introduce the 3D Spatial Reasoning Dataset, comprising 1.74M samples (8.08M QA pairs). 
Unlike the Spatial Understanding dataset, which focuses on qualitative, metric-agnostic spatial concepts (\eg, left, far, inside), this part is metric-grounded and supports flexible output in appropriate units (\eg, cm, inch, m).
Following the TraceSpatial~\cite{zhou2025robotracer} construction pipeline, we propose a data pipeline that progressively integrates 3D scanning and video sources to perform 3D spatial referring, measuring, and tracing.
\textbf{(1) 3D Scanning datasets} want to arm the model with a focused metric-grounded spatial reasoning of indoor scenes.
We thus leverage the richly annotated CA-1M~\cite{lazarow2024cubify} and ScanNet~\cite{dai2017scannet}. 
After fine-grained filtering, similar to the Spatial Understanding part, we construct pseudo-3D scene graphs with more diverse spatial relations, enabled by precise 3D bounding boxes compared to 2D approaches.
Moreover, we generate 3D occupancy maps that encode positions, orientations, and metric distances (\eg, ``35cm right of the toy'') for accurate object-centric spatial trace generation.
\textbf{(2) Manipulation videos} provide spatial traces aligned with the embodied manipulation in tabletop settings.
While 3D scans enable object-centric tracing, they lack physically plausible manipulations for robotics.
Hence, we curate both real (\eg, AgiBot-Beta~\cite{contributors2024agibotworldrepo}, DROID~\cite{khazatsky2024droid}) and simulated (\eg, RoboTwin 2.0~\cite{chen2025robotwin}) tabletop videos. Through a rigorous data cleaning process, such as verifying valid camera poses, coherent task flows, and clean trajectories, we reduce the dataset from 167K to 59K samples for AgiBot-Beta, and from 116K to 24K for DROID.
We further leverage Qwen3-VL~\cite{qwen2.5-vl} to decompose these tasks into subgoals, enabling precise multi-step spatial tracing for single-/dual-arm across $3$ robot configurations.

\vspace{2em}

\subsection{Temporal Prediction Data}

\textbf{Ego-View Planning.}
We construct Ego-View Planning dataset by partially processing the EgoPlan-IT~\cite{chen2024egoplanbenchbenchmarkingmultimodallarge} dataset, which contains 50K automatically generated samples. For each selected task instance, we extract multiple frames from prior actions to represent task progress, and one frame to capture the current viewpoint. To enhance linguistic variety, we use multiple prompt templates that describe the task goal, video context, and current observation. Each question includes the correct next action along with up to three distractor actions randomly sampled from negative examples. This setup supports multimodal instruction tuning with diverse visual and textual input, aimed at improving egocentric task planning performance.

\vspace{0.3em}
\textbf{ShareRobot Planning.}
The ShareRobot dataset~\cite{ji2025robobrain} is a large-scale, fine-grained resource for robotic manipulation, offering multi-dimensional annotations tailored for task planning. Its planning component provides detailed low-level instructions aligned with individual video frames, effectively transforming high-level task descriptions into structured and executable sub-tasks. Each data instance includes precise planning annotations to support accurate and consistent task execution. The dataset comprises 1M QA pairs from 51K instances, spanning 102 diverse scenes across 12 robot embodiments and 107 atomic tasks filtered according to the Open-X-Embodiment taxonomy~\cite{o2024open}. All planning data were meticulously annotated by human experts following the RoboVQA~\cite{sermanet2024robovqa} format, enabling models to learn robust multi-step planning strategies grounded in diverse real-world scenarios. The scale, quality, and diversity of ShareRobot help improve the model's ability to perform fine-grained reasoning and task decomposition in complex embodied environments.

\vspace{0.3em}
\textbf{AGIbot Planning.}
The AgiBot Planning dataset is a large-scale robotics task planning dataset built upon the AgiBot-World~\cite{bu2025agibot} dataset, comprising 9,148 QA pairs across 19 manipulation tasks with 109,378 first-person perspective images. Each sample contains 4-17 consecutive frames documenting task progression with multimodal conversational format. AgiBot-Planning provides step-by-step planning instructions that transform high-level goals into executable sub-tasks. Each data point includes current objectives, historical steps, and required subsequent actions. The dataset covers diverse scenarios from household refrigerator operations to supermarket shopping tasks across different environments. The meticulously crafted annotations use standardized conversational formats, enabling models to learn from varied real-world contexts. Through continuous visual sequences and fine-grained action plans, AgiBot-Planning enhances RoboBrain 2.5’s ability to perform long-horizon task planning and spatial reasoning in complex embodied scenarios.

\vspace{0.3em}
\textbf{Multi-Robot Planning.}
The Multi-Robot Planning dataset is constructed by simulating collaborative task scenarios across three environments—household, supermarket, and restaurant—based on RoboOS~\cite{tan2025roboos, tan2025roboosnext}. Each sample is generated using structured templates that specify a detailed scene graph, robot specifications, and associated tool lists. For every scenario, we design high-level, long-horizon collaborative task goals that require coordination among multiple robots present in the scene, and generate corresponding workflow graphs that decompose the tasks into subtasks with detailed reasoning explanations. Based on these decompositions, we further generate agent-specific robotic tool plans that translate high-level task goals into precise low-level Observation-Action pairs for each subtask. Specifically, we define 1,659 types of multi-robot collaboration tasks across the three environments and produce 44,142 samples using DeepSeek-V3~\cite{deepseekv3}.

\vspace{0.3em}
\textbf{Close-Loop Interaction.}
The Close-Loop Interaction dataset is designed to facilitate advanced embodied reasoning~\cite{zhou2024code}, featuring a large-scale collection of synthesized Observation-Thought-Action (OTA) trajectories that combine first-person visual observations with structured thought tokens. It spans 120 diverse indoor environments—including kitchens, bathrooms, bedrooms, and living rooms—containing over 4,000 interactive objects and receptacles. The dataset is constructed within the AI2Thor~\cite{kolve2017ai2} simulator through a rigorous multi-stage pipeline based on Embodied-Reasoner~\cite{zhang2025embodied}, which includes: (1) crafting task instructions from constrained templates to ensure scene-appropriate validity; (2) deriving key action sequences from an object-affiliation graph encoding functional relationships; and (3) strategically incorporating search actions to emulate realistic exploration. To enrich the depth of reasoning, GPT-4o~\cite{gpt4o} generates detailed thought processes—covering situational analysis, spatial reasoning, self-reflection, task planning, and verification—which are seamlessly integrated between observations and actions, forming coherent reasoning chains that guide models through complex, long-horizon interactive tasks.

\vspace{0.3em}
\textbf{Dense Value Estimation (RoboBrain 2.5 New Feature).} To empower the dense temporal value estimator with robust generalization capabilities, we construct a comprehensive dataset comprising approximately 35 million value estimation samples derived from over 27 million raw frames, and then down-sample to ~3.5M for final training. Following the Dopamine-Reward~\cite{tan2025robo} pipeline, this corpus is meticulously aggregated from three complementary domains, strategically balanced to bridge the gap between physical reality and semantic understanding: 
\textbf{(1) Real-World robot data}, which constitutes the majority ($\sim$60\%) of the training set, integrating diverse datasets such as AGIBot-World~\cite{bu2025agibot}, DROID~\cite{khazatsky2024droid}, and RoboBrain-X~\cite{FlagOpen_RoboBrainX0} to ground the model in physical interaction dynamics across varied environments; 
\textbf{(2) Simulation data} ($\sim$13\%), incorporating benchmarks like LIBERO~\cite{liu2023libero}, RoboCasa~\cite{nasiriany2024robocasa}, and RoboTwin~\cite{chen2025robotwin} to foster strong instruction-following capabilities through high-quality, occlusion-free labels; 
and \textbf{(3) Human-Centric data} ($\sim$26\%), leveraging the massive scale of EgoDex~\cite{hoque2025egodex} to acquire universal object affordance priors independent of robot morphology. Crucially, this heterogeneous mixture spans a wide spectrum of embodiments, ranging from single-arm industrial robots (\eg, Franka Emika Panda) to complex bimanual humanoids (\eg, AGIBot-A2D), preventing overfitting to specific kinematics and ensuring the model focuses on object state changes. We apply the hop-based labeling strategy described in \Cref{subsec:dense_temporal_value_estimation} to this multi-source collection, enabling the model to provide stable, embodiment-invariant progress feedback across a wide spectrum of tasks.


\section{Training Strategy}
\label{sec:trainstrategy}

Similar to RoboBrain 2.0~\cite{team2025robobrain}, RoboBrain 2.5 achieves embodied capabilities (spatial understanding, temporal modeling) through a progressive dual-phase training strategy, as shown in~\Cref{tab:training_setting}. Starting from a robust vision-language foundation, we introduce escalating complexity in embodied supervision, enabling the model to evolve from static perception to dynamic reasoning and actionable planning in real-world environments. Specifically, the training pipeline is divided into two distinct phases: (1) \textit{Foundational Spatiotemporal Learning}, which establishes broad visual semantics, 2D spatial grounding, and open-loop planning capabilities; and (2) \textit{Specific Spatiotemporal Enhancement}, which fine-tunes the model on quantitative 3D spatial reasoning and dense temporal value estimation to ensure precise, metric-aware physical interaction.

\begin{table*}[!ht]
    \centering
    \caption{Detailed configuration for each training stage of the RoboBrain 2.5.}
    \label{tab:training_setting}
    \setlength{\tabcolsep}{12pt}
    \renewcommand{\arraystretch}{1.2}
    \resizebox{0.93\textwidth}{!}{%
    \begin{tabular}{@{}ll|c|c|c|c}
        \toprule
        & & \multicolumn{2}{c|}{\textbf{NVIDIA GPUs}} & \multicolumn{2}{c}{\textbf{Moore Threads GPUs}} \\ \cmidrule(l){3-3} \cmidrule(l){4-4} \cmidrule(l){5-6}
        & & \textbf{Stage-1} & \textbf{Stage-2} & \textbf{Stage-1} & \textbf{Stage-2} \\
        \midrule 
        \multirow{2}{*}{\rotatebox[origin=c]{90}{\small \textit{Data}}}
        & \textbf{Dataset}  & Foundational Learning & Specific Learning & Foundational Learning & Specific Learning \\
        & \#Samples & 8.3 M & 4.1 M & 8.3 M & 4.1 M \\
        \midrule 
        \multirow{2}{*}{\rotatebox[origin=c]{90}{\small \textit{Model}}}
        & \textbf{Trainable Part} & Full Model & Full Model & Full Model & Full Model \\
        & \#Tunable Parameters & 8B & 8B & 8B & 8B \\
        \midrule 
        \multirow{14}{*}{\rotatebox[origin=c]{90}{\small \textit{Training}}}
        & \textbf{Global Batch Size} & 1024 & 1024 & 1024 & 1024 \\
        & \textbf{Tensor Parallelism (TP)} & 2 & 2 & 2 & 2 \\   
        & \textbf{Pipeline Parallelism (PP)} & 2 & 2 & 2 & 2 \\   
        & \textbf{LR: $\{\psi_v^{\text{ViT}}, \phi_v^{\text{LLM}}\}$} & 1$\times 10^{-6}$, 1$\times 10^{-5}$ & 1$\times 10^{-6}$, 1$\times 10^{-5}$ & 1$\times 10^{-6}$, 1$\times 10^{-5}$ & 1$\times 10^{-6}$, 1$\times 10^{-5}$  \\
        & \textbf{Epoch} & 1 & 1 & 1 & 1 \\
        & \textbf{Optimizer} & AdamW & AdamW & AdamW & AdamW \\
        & \textbf{Weight Decay} & 0.1 & 0.1 & 0.1 & 0.0 \\
        & \textbf{Warmup Ratio} & 0.01 & 0.01 & 0.03 & 0.00 \\
        & \textbf{LR Schedule} & Cosine & Cosine & Cosine & Cosine \\
        & \textbf{Max Seq. Length} & 16384 & 16384 & 16384 & 16384 \\
        & \textbf{GPU Nums} & 64 $\times$ 8 & 64 $\times$ 8 & 128 $\times$ 8 & 128 $\times$ 8 \\
        \bottomrule
    \end{tabular}
    }
    \vspace{1mm}
    \vspace{-0.5em}
\end{table*}

\subsection{Stage 1: Foundational Spatiotemporal Learning}
In the first stage, we focus on establishing a robust ``Generalist Brain'' capable of understanding multimodal instructions, grounding objects in 2D space, and mastering high-level planning logic. We utilize the Full Model across \textbf{8.3 million} samples, comprising the \textit{General MLLM Data}, \textit{Spatial Reasoning Data} (excluding metric 3D points/traces), and \textit{Temporal Prediction Data} (Planning and pairwise comparisons).
To ensure stable convergence on this heterogeneous corpus, we employ a standard next-token prediction loss. The primary objectives of this stage are threefold:
\textbf{(1) General Visual Perception:} Leveraging high-quality general data (\eg, Honey-Data-1M) to maintain and enhance the model's general visual-linguistic capabilities. This ensures the model retains a robust understanding of open-world semantics, complex user queries, and diverse visual scenes, serving as a versatile foundation for specific embodied tasks.
\textbf{(2) 2D Grounding \& Qualitative 3D Understanding:} Beyond standard 2D visual grounding and affordance detection, this stage incorporates text-based QA from the 3D Spatial Reasoning dataset. This enables the model to comprehend complex spatial relationships (\eg, spatial relations, occupancy) and qualitative 3D concepts without the burden of precise metric coordinate regression.
\textbf{(3) Planning \& Temporal Logic:} We integrate diverse planning datasets to teach logical task decomposition. Furthermore, we introduce a \textit{Temporal Value Comparison} task derived from the Dense Value Estimation dataset. Instead of predicting absolute values, the model learns to order keyframes temporally (\ie, identifying which frame represents a later state), establishing a preliminary awareness of task progress and state evolution.
This stage yields a model proficient in general perception, logical planning, and qualitative spatiotemporal reasoning, providing a solid initialization for fine-grained training.

\subsection{Stage 2: Specific Spatiotemporal Enhancement}
To bridge the gap between semantic understanding and physical actuation, the second stage introduces Specific Spatiotemporal Enhancement, focusing on precise quantitative reasoning. This stage utilizes approximately \textbf{4.1 million} samples, targeting the newly introduced \textit{Metric 3D Spatial Reasoning} and \textit{Dense Value Estimation} capabilities.
\textbf{(1) Metric-Aware 3D Tracing.} We introduce the specific 3D data focusing on point and trajectory generation to transition the model from qualitative understanding to quantitative perception. This enables the model to predict absolute 3D coordinates, depth-aware traces, and metric distances (\eg, in centimeters), which are critical for precision manipulation tasks.
\textbf{(2) Dense Value Estimation.} We transition from pairwise comparison to explicit \textit{Hop} prediction. The model is trained to act as a robust value function (Critic) by predicting continuous progress values (Hops) frame-by-frame, enabling it to provide fine-grained, closed-loop feedback for policy ranking and error recovery.
\textbf{(3) Anti-Forgetting Strategy.} To prevent the catastrophic forgetting of general capabilities while learning these specialized metric tasks, we adopt a data replay strategy. We randomly sample 15\% of the Stage-1 data and mix it with the Stage-2 specific data. This ensures the model retains its conversation, 2D grounding, and logical planning abilities while mastering fine-grained physical skills for 3D embodied environment.
\section{Infrastructure}

During the training of RoboBrain~2.5, we build upon the infrastructure established in RoboBrain~2.0~\cite{team2025robobrain, ji2025robobrain} while further strengthening and systematizing the core training pipeline. The overall system adopts a multi-dimensional hybrid parallelism strategy, combined with distributed data loading optimizations, and a deeply optimized memory pre-allocation mechanism tailored for multi-modal long-sequence training. These improvements significantly enhance hardware utilization efficiency and overall training throughput.

On the data side, our implementation is based on the Megatron--Energon~\cite{megatron-energon} framework with substantial in-house optimizations. This design enables unified format representation and online mixed training of heterogeneous modalities, including text, single-image, multi-image, and video samples. At the same time, we strictly preserve intra-dataset sample ordering to satisfy the requirements of instruction alignment and temporal consistency. By adopting a customized WebDataset~\cite{webdataset} sample format, the system achieves compatibility with diverse data types while substantially reducing offline preprocessing overhead and improving the flexibility and extensibility of the data pipeline.

\subsection{Hybrid Parallelism}

Multi-modal large models exhibit pronounced heterogeneity in both model architecture and computational characteristics~\cite{liu2024improved}. The visual component typically consists of a relatively lightweight ViT-based encoder (with adapter modules), whereas the language component is dominated by a large-scale decoder-only architecture. Although the visual encoder has a smaller parameter footprint, its computational cost becomes non-trivial when training with a high proportion of visual or video samples.

\vspace{0.3em}

To address this architectural heterogeneity, we leverage the heterogeneous training experience accumulated in our in-house distributed framework, FlagScale~\cite{flagscale}, and adopt an \emph{uneven pipeline parallelism} strategy~\cite{megatron}. Specifically, the ViT module is placed at the front of the model, and the number of language layers assigned to the first pipeline stage is reduced accordingly. This design balances computational load across pipeline stages, mitigates pipeline bubbles, and improves overall pipeline efficiency.

\subsection{Dynamic pre-Allocated Memory}

In RoboBrain~2.5 training, sequence lengths vary significantly across samples. Combined with PyTorch's default CUDA caching memory allocator, this dynamic-shape workload often leads to severe GPU memory fragmentation and, in extreme cases, out-of-memory (OOM) failures. A common workaround is to invoke \texttt{torch.cuda.empty\_cache()}~\cite{pytorch-memory} before each iteration; however, this approach disrupts memory reuse and substantially degrades training performance.

\vspace{0.3em}

To resolve this issue, we conduct an in-depth analysis of CUDA memory allocation and reuse behavior and propose a \emph{dynamic unified padding strategy based on dual data streams}.

\vspace{0.3em}

\begin{itemize}
\item Before training begins, the maximum sequence length observed in the training set is collected;
\item In the first training iteration, all samples are padded to this maximum length, enabling one-time memory pre-allocation during initialization;
\item In subsequent iterations, tensors reuse the pre-allocated memory, effectively suppressing memory fragmentation;
\item Only when the visual token length exceeds the current maximum does the system trigger a full cache cleanup and re-pad samples to the new maximum length.
\end{itemize}

\vspace{0.3em}

This strategy strikes a practical balance between memory efficiency and training performance, providing both stability and high throughput in large-scale multi-modal long-sequence training scenarios.

\subsection{Cross-Accelerator Training and Inference}

Leveraging FlagScale's distributed training capabilities on heterogeneous accelerator clusters, together with VLM-specific kernel and communication optimizations, we successfully complete end-to-end training of RoboBrain~2.5 on a thousand-device cluster composed of \emph{non-NVIDIA accelerators}. The resulting loss convergence behavior closely matches that observed on NVIDIA platforms, with the final convergence gap controlled within 0.62\%.

\vspace{0.3em}

Furthermore, the trained checkpoints are seamlessly migrated to NVIDIA-based platforms for downstream evaluation. Across a range of mainstream benchmarks, the resulting performance remains highly consistent with models trained natively on NVIDIA hardware. This RoboBrain~2.5 case study demonstrates that FlagOS/FlagScale's cross-accelerator training and inference capabilities have matured to a level that is \emph{reliable, practical, and production-ready} for large-scale multi-modal model training.

\newcommand{\gemini}{Gemini 2.5 Pro}
\newcommand{\openaio}{OpenAI o1}
\newcommand{\claude}{Claude 3.7 Sonnet}
\newcommand{\openaigpt}{OpenAI GPT-4o}
\newcommand{\qwen}{Qwen 2.5-VL 72B}
\newcommand{\openaicua}{OpenAI CUA}

\section{Evaluation Results}
\label{sec:evaluation}

We conducted a comprehensive evaluation of RoboBrain-2.5, significantly expanding the assessment scope of its predecessor to include 3D quantitative spatial reasoning and fine-grained temporal value estimation. To ensure consistency and rigor, we continued to employ FlagEvalMM~\cite{FlagEvalMM}, our flexible framework for systematic multimodal model assessment. Notably, to demonstrate the cross-platform robustness of our training infrastructure, we report performance for RoboBrain-2.5 variants trained on two distinct hardware backends: \textbf{NVIDIA (NV) GPUs} and \textbf{Moore-Threads (MTT) GPUs}.

\vspace{0.3em}

Evaluations on spatial reasoning benchmarks, which now encompass both foundational 2D tasks (\eg, CV-Bench~\cite{tong2024cambrian}, RoboSpatial~\cite{song2025robospatial}) and advanced 3D quantitative measurement (\eg, MSMU ~\cite{chen2025sd}, TraceSpatial~\cite{zhou2025robotracer}, VABench-V~\cite{yuan2025embodied}), are presented in \Cref{subsec:2d_spatial_capability} and \Cref{subsec:3d_spatial_capability}.  Furthermore, we introduce a new dimension of evaluation for temporal value estimation in \Cref{subsec:temporal_capability}, leveraging the General Process Reward Modeling (GPRM) paradigm from Robo-Dopamine~\cite{tan2025robo}. We assess the model's ability to perceive manipulation progress across diverse data sources, organized into \textbf{Real-Bench} (real-world robot data including AgiBot ~\cite{bu2025agibot}, DROID~\cite{khazatsky2024droid}, and Galaxea~\cite{jiang2025galaxea}), \textbf{Sim-Bench} (simulation environments like Libero ~\cite{liu2023libero}and RoboCasa~\cite{nasiriany2024robocasa}), and \textbf{Human-Bench} (human manipulation videos from EgoDex~\cite{hoque2025egodex}). Qualitative examples are provided in \Cref{sec:app:qualitative}.

\vspace{2em}
\subsection{2D Spatial Reasoning Capability}
\label{subsec:2d_spatial_capability} 

We first evaluate RoboBrain-2.5 on five representative \textbf{2D spatial reasoning} benchmarks: \textbf{CV-Bench}~\cite{tong2024cambrian}, \textbf{CrossPoint}~\cite{wang2025towards}, \textbf{RoboSpatial}~\cite{song2025robospatial}, \textbf{RefSpatial}~\cite{zhou2025roborefer}, and \textbf{EmbSpatial}~\cite{du2024embspatial}. Results are summarized in~\Cref{tab:2DSpatial}. Overall, the RoboBrain-2.5 variants trained on the \textbf{NVIDIA GPU Platform} and \textbf{Moore-Threads (MTT) GPU Platform} achieve same average scores of \textbf{75.82}. Both deliver substantial improvements over general-purpose and embodied baselines.

\vspace{0.3em}

\begin{table*}[!t]
    \centering
    \caption{\textbf{Performance on 2D spatial reasoning benchmarks.}
    The best results are highlighted in \textbf{bold}, while the second-best results are \underline{underlined}. Results marked with * are sourced from their technical reports.}
    \vspace{-0.5em}
    \resizebox{0.98\textwidth}{!}{
    \begin{tabular}{l|ccccc|c}
        \toprule
        \multicolumn{1}{l|}{\multirow{2}{*}{\textbf{Models / Metrics}}} &
        \multicolumn{5}{c|}{\textbf{2D Spatial Reasoning}} &
        \textbf{AVG} \\
        \cmidrule(lr){2-6} \cmidrule(lr){7-7}
        & \textbf{CV-Bench}  & \textbf{CrossPoint} & \textbf{RoboSpatial} & \textbf{RefSpatial}
        & \textbf{EmbSpatial} & \textbf{All $\uparrow$} \\
        \midrule

        \rowcolor[HTML]{F2F2F2} \multicolumn{7}{l}{\textbf{General Baselines}} \\ \midrule
        Gemini-3-Pro-Preview~\cite{gemini3pro}&
        92.00$^*$ & 38.60 & 57.96 & \textbf{65.50}$^*$ &
        76.62 &
        \underline{66.14} \\

        GPT-5.2~\cite{gpt5.2}&
        86.84 & 33.00 & 43.78 & 15.00 &
        68.02 &
        49.33 \\

        Qwen3-VL-8B-Inst.~\cite{qwen3vl}&
        92.89 & 28.40 & 66.90$^*$ & 54.20$^*$ &
        \textbf{78.50}$^*$ &
        64.18 \\

        \midrule
        \rowcolor[HTML]{F2F2F2} \multicolumn{7}{l}{\textbf{Embodied Baselines}} \\ \midrule
        RoboBrain-2.0 (7B)~\cite{team2025robobrain} &
        85.75$^*$ & 26.00 & 54.23$^*$ & 32.50$^*$ &
        76.32$^*$ &
        54.96 \\

        Mimo-Embodied (7B)~\cite{hao2025mimo}&
        88.82$^*$ & 20.02 & 61.76$^*$ & 48.00$^*$ &
        76.24$^*$ &
        58.97 \\

        \rowcolor[HTML]{DAEFF9} RoboBrain-2.5 (8B) \textbf{NV} &
        \textbf{94.58} & \underline{75.40} & \textbf{73.03} & \underline{60.50} &
        75.58 &
        \textbf{75.82} \\

        \rowcolor[HTML]{DAEFF9} RoboBrain-2.5 (8B) \textbf{MTT} &
        \underline{93.90} & \textbf{76.30} & \underline{73.00} & 59.00 &
        \underline{76.92} &
        \textbf{75.82} \\

        \bottomrule
    \end{tabular}
    }
    \label{tab:2DSpatial}
\vspace{-0.5em}
\end{table*}

\vspace{+1mm}

\begin{itemize}

\item{\textbf{CV-Bench~\cite{tong2024cambrian}.}}
CV-Bench assesses vision-centric spatial understanding and visual processing via repurposed 2D/3D vision tasks. RoboBrain-2.5 (8B) trained on NVIDIA achieves the best accuracy of \textbf{94.58}, with the MTT variant closely following at 93.90. Both consistently outperform strong general baselines such as Qwen3-VL-8B-Inst. (92.89), Gemini-3-Pro-Preview (92.00), and GPT-5.2 (86.84), as well as embodied baselines including RoboBrain-2.0 (7B) (85.75) and Mimo-Embodied (7B) (88.82), indicating a clear gain in foundational 2D spatial perception.

\vspace{+1mm}

\item{\textbf{CrossPoint~\cite{wang2025towards}.}}
CrossPoint-Bench evaluates cross-view point correspondence, requiring fine-grained point-level matching across different viewpoints. RoboBrain-2.5 demonstrates a decisive advantage, achieving \textbf{76.30} (MTT) and 75.40 (NVIDIA), which substantially surpasses all evaluated baselines, including Gemini-3-Pro-Preview (38.60), GPT-5.2 (33.00), Qwen3-VL-8B-Inst. (28.40), RoboBrain-2.0 (7B) (26.00), and Mimo-Embodied (7B) (20.02). This highlights the model’s strong capability in transitioning from coarse spatial judgment to actionable, coordinate-level correspondence.

\vspace{+1mm}

\item{\textbf{RoboSpatial~\cite{song2025robospatial}.}}
RoboSpatial measures spatial reasoning in robotics-oriented environments, emphasizing egocentric understanding, reference frames, and interaction-relevant spatial relations. RoboBrain-2.5 achieves the best scores of \textbf{73.03} (NVIDIA) and 73.00 (MTT), outperforming Qwen3-VL-8B-Inst. (66.90) and Gemini-3-Pro-Preview (57.96), as well as embodied baselines like Mimo-Embodied (7B) (61.76) and RoboBrain-2.0 (7B) (54.23). The consistent gains suggest improved spatial grounding for robot-centric perception and interaction.

\vspace{+1mm}

\item{\textbf{RefSpatial~\cite{zhou2025roborefer}.}}
RefSpatial evaluates spatial referring under complex spatial constraints, demanding precise grounding with multi-step spatial reasoning. RoboBrain-2.5 achieves strong results of 60.50 (NVIDIA) and 59.00 (MTT), substantially exceeding Qwen3-VL-8B-Inst. (54.20), Mimo-Embodied (7B) (48.00), RoboBrain-2.0 (7B) (32.50), and GPT-5.2 (15.00), while remaining competitive with the best-performing general baseline (Gemini-3-Pro-Preview, \textbf{65.50}). This indicates robust spatial referring performance in cluttered, instruction-conditioned settings.

\vspace{+1mm}

\item{\textbf{EmbSpatial~\cite{du2024embspatial}.}}
EmbSpatial-Bench assesses embodied spatial understanding from an egocentric perspective. RoboBrain-2.5 attains competitive performance with 76.92 (MTT) and 75.58 (NVIDIA), closely matching Gemini-3-Pro-Preview (76.62) and surpassing GPT-5.2 (68.02), while approaching the strongest baseline Qwen3-VL-8B-Inst. (\textbf{78.50}). These results suggest that RoboBrain-2.5 achieves strong generalization in embodied spatial relations, with minimal sensitivity to the training hardware backend.

\end{itemize}

\subsection{3D Spatial Reasoning Capability}
\label{subsec:3d_spatial_capability}

We further evaluate RoboBrain-2.5 on five \textbf{3D spatial reasoning} benchmarks that stress \emph{metric-grounded} and \emph{trajectory-aware} understanding: \textbf{MSMU}~\cite{chen2025sd}, \textbf{Q-Spatial}~\cite{liao2024reasoning}, \textbf{TraceSpatial}~\cite{zhou2025robotracer}, \textbf{VABench-V}~\cite{yuan2025embodied}, and \textbf{ShareRobot-Bench}~\cite{ji2025robobrain}. Results are summarized in~\Cref{tab:3DSpatial}. Unless otherwise noted, higher is better; specifically for \textbf{VABench-V} and \textbf{ShareRobot-Bench}, we report distance-based metrics where lower indicates better performance.
\begin{table*}[!t]
    \centering
    \caption{\textbf{Performance on five 3D spatial reasoning benchmarks.}
    For \textbf{TraceSpatial}, we further report fine-grained 3D metrics including \textbf{3D Start}, \textbf{3D End}, and \textbf{Success} for the detailed trace evaluation.
    The best results among different models are highlighted in \textbf{bold}, while the second-best results are \underline{underlined}.}
    \vspace{-0.5em}
    \resizebox{0.98\textwidth}{!}{
    \begin{tabular}{l|cc|ccc|cc}
        \toprule
        \multicolumn{1}{l|}{\multirow{3}{*}{\textbf{Models / Metrics}}} &
        \multicolumn{7}{c}{\textbf{3D Spatial Reasoning}} \\
        \cmidrule(lr){2-8}
        & \multirow{3}{*}{\textbf{MSMU $\uparrow$}} & \multirow{3}{*}{\textbf{Q-Spatial $\uparrow$}} &
        \multicolumn{3}{c|}{\textbf{TraceSpatial $\uparrow$}} &
        \multirow{3}{*}{\textbf{VABench-V $\downarrow$}} & \multirow{3}{*}{\textbf{ShareRobot-T $\downarrow$}} \\
        \cmidrule(lr){4-6}
        &  &  & \textbf{3D Start} & \textbf{3D End} & \textbf{Success} &  &  \\
        \midrule

        \rowcolor[HTML]{F2F2F2} \multicolumn{8}{l}{\textbf{General Baselines}} \\ \midrule
        Gemini-3-Pro-Preview~\cite{gemini3pro} & 59.44 & \textbf{81.37} & 19 & 25 & 7  & 0.1705 & 0.1899 \\
        GPT-5.2~\cite{gpt5.2}  & 57.96 & 69.16          & 3     & 8     & 0  & 0.1962 & 0.2379 \\
        Qwen3-VL-8B-Inst.~\cite{qwen3vl}& 43.48 & 70.74     & 30    & 20    & 6  & 0.1979 & 0.2347 \\

        \midrule
        \rowcolor[HTML]{F2F2F2} \multicolumn{8}{l}{\textbf{Embodied Baselines}} \\ \midrule
        RoboBrain-2.0 (7B)~\cite{team2025robobrain}& 55.01 & 63.37 & -- & -- & -- & -- & 0.1240 \\
        Mimo-Embodied (7B)~\cite{hao2025mimo}& 46.36 & 65.42 & -- & -- & -- & 0.6970 & 0.6351 \\

        \rowcolor[HTML]{DAEFF9} RoboBrain-2.5 (8B) \textbf{NV} &
        \textbf{64.17} & 73.53 & \textbf{83} & \underline{63} & \textbf{44} & \underline{0.1281} & \textbf{0.1164} \\

        \rowcolor[HTML]{DAEFF9} RoboBrain-2.5 (8B) \textbf{MTT} &
        \underline{61.66} & \underline{78.31} & \underline{80} & \textbf{65} & \underline{36} & \textbf{0.1189} & \underline{0.1171} \\

        \bottomrule
    \end{tabular}
    }
    \label{tab:3DSpatial}
\vspace{-0.5em}
\end{table*}

\vspace{+1mm}

\begin{itemize}

\item{\textbf{MSMU~\cite{chen2025sd}.}}
MSMU evaluates quantitative 3D spatial measuring and understanding with precise numerical annotations. RoboBrain-2.5 achieves the best performance, with \textbf{64.17} (NVIDIA) and 61.66 (MTT), surpassing strong general baselines such as Gemini-3-Pro-Preview (59.44) and GPT-5.2 (57.96), as well as embodied baselines like RoboBrain-2.0 (55.01) and Mimo-Embodied (46.36), indicating substantially improved metric-grounded perception.

\vspace{+1mm}

\item{\textbf{Q-Spatial~\cite{liao2024reasoning}.}}
Q-Spatial Benchmark assesses quantitative reasoning about object sizes and distances in images. RoboBrain-2.5 (MTT) achieves a strong score of \underline{78.31}, outperforming Qwen3-VL-8B-Inst. (70.74), GPT-5.2 (69.16), RoboBrain-2.0 (63.37), and Mimo-Embodied (65.42), while remaining competitive with the best-performing general baseline (Gemini-3-Pro-Preview, \textbf{81.37}). This demonstrates robust quantitative spatial reasoning without specialized test-time prompting.

\vspace{+1mm}

\item{\textbf{TraceSpatial~\cite{zhou2025robotracer}.}}\
TraceSpatial-Bench evaluates multi-step, metric-grounded \emph{spatial tracing} in cluttered 3D scenes, where a prediction is considered successful only if the trajectory satisfies correct start/end spatial constraints and remains collision-free.
We report three fine-grained 3D metrics: \textbf{3D Start} measures \emph{grasp success} (whether the predicted start point is sufficiently close to the target object point cloud), \textbf{3D End} measures \emph{placement success} (whether the predicted end point falls inside/near the destination object’s 3D bounding box), and \textbf{Success} measures the final \emph{spatial trace success} by jointly considering grasp success, placement success, and collision checking along the trace~\cite{zhou2025robotracer}.

\vspace{+1mm}

\item{\textbf{VABench-V~\cite{yuan2025embodied}.}}
VABench-V evaluates visual trace generation from natural language instructions with distance-based metrics (RMSE) (lower is better). RoboBrain-2.5 achieves a clear SOTA with a lowest error of \textbf{0.1189} (MTT) and a close second-best of \underline{0.1281} (NVIDIA), substantially improving over Gemini-3-Pro-Preview (0.1705), GPT-5.2 (0.1962), and Qwen3-VL-8B-Inst. (0.1979), demonstrating accurate fine-grained waypoint generation.

\vspace{+1mm}

\item{\textbf{ShareRobot-T~\cite{ji2025robobrain}.}}
ShareRobot-Traj Benchmark assesses robot-centric spatial grounding for interaction and trajectory-related prediction (RMSE), where lower distance indicates better performance. RoboBrain-2.5 attains the best results with \textbf{0.1164} (NVIDIA) and a close second of \underline{0.1171} (MTT), improving over RoboBrain-2.0 (0.1240) and strongly outperforming general baselines such as Gemini-3-Pro-Preview (0.1899) and GPT-5.2 (0.2379), reflecting more precise interaction-relevant spatial outputs.

\end{itemize}

\subsection{Temporal Value Estimation}
\label{subsec:temporal_capability}
To evaluate \emph{fine-grained temporal value estimation} for manipulation progress, we follow the General Process Reward Modeling (GPRM) paradigm in Robo-Dopamine~\cite{tan2025robo}. Concretely, the model is prompted with a task instruction and conditioned on multi-view images of the \emph{initial} and \emph{goal} states, together with paired multi-view observations of the \emph{BEFORE} and \emph{AFTER} states, and predicts a discretized relative progress/regress hop as a value signal~\cite{tan2025robo}.
We evaluate temporal ordering robustness via two rank-correlation metrics: \textbf{Forward VOC} (\textbf{VOC$^{+}$}) computed on the original temporal direction, and \textbf{Reverse VOC} (\textbf{VOC$^{-}$}) computed by \emph{time-reversing} the video and re-evaluating the model (i.e., the predicted value should consistently invert with the reversed temporal order). We report VOC$^{+}$ / VOC$^{-}$ (both $\uparrow$) on six data sources spanning real-robot, simulation, and human egocentric videos: AgiBot~\cite{bu2025agibot}, DROID~\cite{khazatsky2024droid}, Galaxea~\cite{jiang2025galaxea}, EgoDex~\cite{hoque2025egodex}, LIBERO~\cite{liu2023libero}, and RoboCasa~\cite{nasiriany2024robocasa}. Results are summarized in~\Cref{tab:TemporalValue}.

\vspace{0.3em}

\begin{table*}[!t]
    \centering
    \caption{\textbf{Temporal value estimation on six testsets.} We report \textbf{VOC$^{+}$ / VOC$^{-}$} (both $\uparrow$), where VOC$^{-}$ is computed by reversing the video and re-evaluating the model. The best results among different models are highlighted in \textbf{bold}, while the second-best results are \underline{underlined}.}
    \vspace{-0.5em}
    \resizebox{0.95\textwidth}{!}{
    \begin{tabular}{l|cccccc}
        \toprule
        \multicolumn{1}{l|}{\multirow{2}{*}{\textbf{Models / Metrics}}} &
        \multicolumn{6}{c}{\textbf{VOC$^{+}$ / VOC$^{-}$} ($\uparrow$)} \\
        \cmidrule(lr){2-7}
        & \textbf{AgiBot} & \textbf{DROID} & \textbf{Galaxea} & \textbf{EgoDex} & \textbf{LIBERO} & \textbf{RoboCasa} \\
        \midrule
        \rowcolor[HTML]{F2F2F2} \multicolumn{7}{l}{\textbf{General Baselines}} \\ \midrule
        Gemini-3-Pro-Preview~\cite{gemini3pro}& 81.36 / 58.70 & 90.57 / 44.15 & 88.86 / 35.34 & \underline{80.48} / 50.15 & 98.42 / 76.31 & 67.89 / 34.28 \\

        GPT-5.2~\cite{gpt5.2}& \textbf{90.02} / 15.91 & \underline{91.45} / 15.29 & 88.76 / 10.03 & 78.12 / 22.79 & 96.97 / 19.19 & 77.91 / 10.71 \\
        
        Qwen3-VL-8B-Inst.~\cite{qwen3vl}& 82.50 / 5.32 & 81.33 / 10.37 & 79.98 / 5.51 & 63.85 / 12.82 & 72.31 / 22.07 & 59.11 / -0.03 \\
        \midrule
        \rowcolor[HTML]{F2F2F2} \multicolumn{7}{l}{\textbf{Embodied Models}} \\ \midrule
        \rowcolor[HTML]{DAEFF9} RoboBrain-2.5 (8B) \textbf{NV} & 83.08 / \textbf{88.58} & 90.82 / \textbf{90.07} & \underline{93.38} / \textbf{95.79} & 79.14 / \textbf{84.99} & \textbf{98.97} / \textbf{98.94} & \underline{98.47} / \underline{98.75} \\
        \rowcolor[HTML]{DAEFF9} RoboBrain-2.5 (8B) \textbf{MTT} & \underline{87.36} / \underline{87.48} & \textbf{93.67} / \underline{89.26} & \textbf{94.58} / \underline{94.54} & \textbf{80.67} / \underline{81.12} & \underline{98.88} / \underline{98.91} & \textbf{98.54} / \textbf{99.58} \\
        \bottomrule
    \end{tabular}
    }
    \label{tab:TemporalValue}
\vspace{-0.5em}
\end{table*}

\vspace{+1mm}

\begin{itemize}

\item{\textbf{AgiBot~\cite{bu2025agibot}.}}
On AgiBot, the \textbf{RoboBrain-2.5} variants demonstrate strong and balanced performance. Specifically, the model trained on Moore-Threads (MTT) achieves \textbf{87.36 / 87.48}, ranking second in Forward VOC while maintaining high consistency. In contrast, while generalist VLMs like GPT-5.2 achieve a higher Forward VOC (\textbf{90.02}), they exhibit substantially lower Reverse VOC (15.91), indicating a lack of robust bidirectional temporal understanding compared to our embodied models.

\vspace{+1mm}

\item{\textbf{DROID~\cite{khazatsky2024droid}.}}
On DROID, \textbf{RoboBrain-2.5 (MTT)} attains a clear lead with \textbf{93.67 / 89.26}, substantially improving over all baselines. While GPT-5.2 achieves the second-best Forward VOC (\underline{91.45}), its Reverse VOC drops sharply to 15.29. Similarly, Gemini-3-Pro-Preview shows a large gap between forward (90.57) and reverse (44.15) performance, again highlighting the benefit of RoboBrain-2.5's step-aware progress supervision.

\vspace{+1mm}

\item{\textbf{Galaxea~\cite{FlagOpen_RoboBrainX0}.}}
On Galaxea, both RoboBrain-2.5 variants perform exceptionally well, with the MTT variant reaching \textbf{94.58 / 94.54} and the NVIDIA variant closely following at \underline{93.38} / 95.79. General baselines show significantly weaker Reverse VOC (e.g., Gemini-3-Pro-Preview at 35.34 and GPT-5.2 at 10.03), suggesting that high forward correlation alone is insufficient without robust time-reversal behavior.

\vspace{+1mm}

\item{\textbf{EgoDex~\cite{hoque2025egodex}.}}
On EgoDex (human egocentric manipulation videos), \textbf{RoboBrain-2.5 (MTT)} achieves the best comprehensive result (\textbf{80.67 / 81.12}). While Gemini-3-Pro-Preview shows competitive Forward VOC (\underline{80.48}), its Reverse VOC (50.15) is significantly lower. This indicates that RoboBrain-2.5 generalizes better to human-centric temporal cues while maintaining logical consistency across temporal directions.

\vspace{+1mm}

\item{\textbf{LIBERO~\cite{liu2023libero}.}}
On LIBERO, both RoboBrain-2.5 models achieve near-ceiling performance, with the NVIDIA variant reaching \textbf{98.97 / 98.94} and the MTT variant close behind at \underline{98.88} / 98.91. Although general baselines show relatively high Forward VOC (e.g., Gemini-3-Pro-Preview at 98.42), their lower Reverse VOC (76.31 or below) reinforces that bidirectional temporal consistency is a stricter criterion for progress-aware value modeling.

\vspace{+1mm}

\item{\textbf{RoboCasa~\cite{nasiriany2024robocasa}.}}
On RoboCasa, \textbf{RoboBrain-2.5 (MTT)} achieves the best performance (\textbf{98.54 / 99.58}), followed closely by the NVIDIA variant. Compared to general baselines (e.g., GPT-5.2 at 77.91 / 10.71), the RoboBrain-2.5 models demonstrate markedly stronger robustness under time reversal, consistent with the design goal of providing reliable, step-aware progress signals for manipulation~\cite{tan2025robo}.

\end{itemize}

\section{Conclusion and Future Works}
\label{sec:conclusion}

In this work, we introduced \textbf{RoboBrain-2.5}, a next-generation embodied AI foundation model that significantly bridges the gap between high-level semantic reasoning and low-level physical interaction. By addressing the fundamental limitations of prior generalist models—specifically the lack of metric-grounded spatial precision and the absence of dense temporal supervision—RoboBrain-2.5 achieves a comprehensive upgrade in embodied capabilities. Our contributions are established through two core pillars. First, we proposed \textbf{Precise 3D Spatial Reasoning}, moving beyond 2D pixel-relative grounding to depth-aware coordinate prediction. By utilizing a decoupled $(u, v, d)$ representation and training on high-quality 3D spatial data, the model learns to interpret absolute metric constraints and generate collision-free, trajectory-level manipulation traces. Second, we introduced \textbf{Dense Temporal Value Estimation}, a mechanism that provides fine-grained, step-aware progress and regress feedback. This capability, powered by a hop-based labeling strategy and multi-perspective fusion, enables the model to serve as a robust general-purpose reward function resilient to viewpoint variations. Furthermore, we demonstrated the scalability of our approach through a robust infrastructure capable of cross-accelerator training on both NVIDIA and Moore Threads GPUs. Extensive evaluations confirm that RoboBrain-2.5 sets a new state-of-the-art on both spatial reasoning and temporal value estimation tasks.
In future research, we plan to expand the capabilities and efficiency of the RoboBrain model series in four primary directions:

\vspace{0.3em}

\begin{itemize}
    \item \textbf{Unified Generation and Understanding Paradigm:} We aim to evolve RoboBrain into a unified architecture that integrates both spatiotemporal understanding and generative capabilities. By incorporating image and video prediction (i.e., next-stage prediction), the model will serve as an embodied world model. This will enable agents to simulate action outcomes in their ``mind'' before execution, significantly enhancing planning safety and robustness in complex environments.

    \vspace{0.3em}
    
    \item \textbf{Deployment on Mobile Manipulation and Humanoids:} We will extensively validate and deploy our models on diverse real-world platforms, including mobile manipulators and humanoid robots~\cite{chen2025ac, li2025language, li2025robomirror, li2025you, li2024lamp}. Our focus will be on leveraging \textit{Precise 3D Spatial Reasoning} to achieve training-free manipulation generalization, while utilizing \textit{Dense Temporal Value Estimation} as a high-fidelity reward signal to drive efficient Reinforcement Learning (RL) in the physical world.

    \vspace{0.3em}

    \item \textbf{Scalable Model Family and Specialized Variants:} To accommodate varying computational constraints and latency requirements, we plan to release a comprehensive series of models with different parameter scales. This includes lightweight versions optimized for edge-device deployment and high-frequency inference, as well as decoupling the architecture into distinct ``Instruction'' (fast execution) and ``Thinking'' (slow reasoning) versions to balance response speed with reasoning depth.

    \vspace{0.3em}

    \item \textbf{Self-Evolving Data Engine:} We intend to establish a closed-loop data engine where RoboBrain 2.5 acts as a verifier for its own data. By utilizing the dense value estimator to automatically filter and annotate large-scale uncurated videos, the model can iteratively improve itself through self-supervised learning, creating a flywheel effect for continuous capability enhancement.
\end{itemize}

\clearpage


\bibliographystyle{plainnat}
\bibliography{main}

\begin{thebibliography}{87}
\providecommand{\natexlab}[1]{#1}
\providecommand{\url}[1]{\texttt{#1}}
\expandafter\ifx\csname urlstyle\endcsname\relax
  \providecommand{\doi}[1]{doi: #1}\else
  \providecommand{\doi}{doi: \begingroup \urlstyle{rm}\Url}\fi

\bibitem[Abdolmaleki et~al.(2025)Abdolmaleki, Abeyruwan, Ainslie, Alayrac, Arenas, Balakrishna, Batchelor, Bewley, Bingham, Bloesch, et~al.]{abdolmaleki2025gemini}
Abbas Abdolmaleki, Saminda Abeyruwan, Joshua Ainslie, Jean-Baptiste Alayrac, Montserrat~Gonzalez Arenas, Ashwin Balakrishna, Nathan Batchelor, Alex Bewley, Jeff Bingham, Michael Bloesch, et~al.
\newblock Gemini robotics 1.5: Pushing the frontier of generalist robots with advanced embodied reasoning, thinking, and motion transfer.
\newblock \emph{arXiv preprint arXiv:2510.03342}, 2025.

\bibitem[Ahn et~al.(2024)Ahn, Dwibedi, Finn, Arenas, Gopalakrishnan, Hausman, Ichter, Irpan, Joshi, Julian, et~al.]{ahn2024autort}
Michael Ahn, Debidatta Dwibedi, Chelsea Finn, Montse~Gonzalez Arenas, Keerthana Gopalakrishnan, Karol Hausman, Brian Ichter, Alex Irpan, Nikhil Joshi, Ryan Julian, et~al.
\newblock Autort: Embodied foundation models for large scale orchestration of robotic agents.
\newblock \emph{arXiv preprint arXiv:2401.12963}, 2024.

\bibitem[Alakuijala et~al.(2024)Alakuijala, McLean, Woungang, Farsad, Kaski, Marttinen, and Yuan]{alakuijala2024video}
Minttu Alakuijala, Reginald McLean, Isaac Woungang, Nariman Farsad, Samuel Kaski, Pekka Marttinen, and Kai Yuan.
\newblock Video-language critic: Transferable reward functions for language-conditioned robotics.
\newblock \emph{arXiv preprint arXiv:2405.19988}, 2024.

\bibitem[Alex~Aizman(2020)]{webdataset}
Thomas~Breuel Alex~Aizman, Gavin~Maltby.
\newblock Webdataset: High-performance data loading for deep learning, 2020.
\newblock URL \url{https://webdataset.github.io/webdataset/}.

\bibitem[An et~al.(2025)An, Xie, Yang, Zhang, Zhao, Cheng, Wang, Xu, Chen, Wu, et~al.]{an2025llava}
Xiang An, Yin Xie, Kaicheng Yang, Wenkang Zhang, Xiuwei Zhao, Zheng Cheng, Yirui Wang, Songcen Xu, Changrui Chen, Chunsheng Wu, et~al.
\newblock Llava-onevision-1.5: Fully open framework for democratized multimodal training.
\newblock \emph{arXiv preprint arXiv:2509.23661}, 2025.

\bibitem[Azuma et~al.(2022)Azuma, Miyanishi, Kurita, and Kawanabe]{scanqa}
Daichi Azuma, Taiki Miyanishi, Shuhei Kurita, and Motoaki Kawanabe.
\newblock Scanqa: 3d question answering for spatial scene understanding.
\newblock In \emph{CVPR}, pages 19129--19139, 2022.

\bibitem[Azzolini et~al.(2025)Azzolini, Brandon, Chattopadhyay, Chen, Chu, Cui, Diamond, Ding, Ferroni, Govindaraju, et~al.]{cosmos-reason1}
Alisson Azzolini, Hannah Brandon, Prithvijit Chattopadhyay, Huayu Chen, Jinju Chu, Yin Cui, Jenna Diamond, Yifan Ding, Francesco Ferroni, Rama Govindaraju, et~al.
\newblock Cosmos-reason1: From physical common sense to embodied reasoning.
\newblock \emph{arXiv preprint arXiv:2503.15558}, 2025.

\bibitem[Bai et~al.(2025{\natexlab{a}})Bai, Cai, Chen, Chen, Chen, Cheng, Deng, Ding, Gao, Ge, Ge, Guo, Huang, Huang, Huang, Hui, Jiang, Li, Li, Li, Li, Lin, Lin, Liu, Liu, Liu, Liu, Liu, Liu, Lu, Luo, Lv, Men, Meng, Ren, Ren, Song, Sun, Tang, Tu, Wan, Wang, Wang, Wang, Wang, Xie, Xu, Xu, Xu, Yang, Yang, Yang, Yang, Yu, Zhang, Zhang, Zhang, Zheng, Zhong, Zhou, Zhou, Zhou, Zhu, and Zhu]{qwen3vl}
Shuai Bai, Yuxuan Cai, Ruizhe Chen, Keqin Chen, Xionghui Chen, Zesen Cheng, Lianghao Deng, Wei Ding, Chang Gao, Chunjiang Ge, Wenbin Ge, Zhifang Guo, Qidong Huang, Jie Huang, Fei Huang, Binyuan Hui, Shutong Jiang, Zhaohai Li, Mingsheng Li, Mei Li, Kaixin Li, Zicheng Lin, Junyang Lin, Xuejing Liu, Jiawei Liu, Chenglong Liu, Yang Liu, Dayiheng Liu, Shixuan Liu, Dunjie Lu, Ruilin Luo, Chenxu Lv, Rui Men, Lingchen Meng, Xuancheng Ren, Xingzhang Ren, Sibo Song, Yuchong Sun, Jun Tang, Jianhong Tu, Jianqiang Wan, Peng Wang, Pengfei Wang, Qiuyue Wang, Yuxuan Wang, Tianbao Xie, Yiheng Xu, Haiyang Xu, Jin Xu, Zhibo Yang, Mingkun Yang, Jianxin Yang, An~Yang, Bowen Yu, Fei Zhang, Hang Zhang, Xi~Zhang, Bo~Zheng, Humen Zhong, Jingren Zhou, Fan Zhou, Jing Zhou, Yuanzhi Zhu, and Ke~Zhu.
\newblock Qwen3-vl technical report, 2025{\natexlab{a}}.
\newblock URL \url{https://arxiv.org/abs/2511.21631}.

\bibitem[Bai et~al.(2025{\natexlab{b}})Bai, Chen, Liu, Wang, Ge, Song, Dang, Wang, Wang, Tang, et~al.]{qwen25vl}
Shuai Bai, Keqin Chen, Xuejing Liu, Jialin Wang, Wenbin Ge, Sibo Song, Kai Dang, Peng Wang, Shijie Wang, Jun Tang, et~al.
\newblock Qwen2. 5-vl technical report.
\newblock \emph{arXiv preprint arXiv:2502.13923}, 2025{\natexlab{b}}.

\bibitem[Bai et~al.(2025{\natexlab{c}})Bai, Song, Chen, Ji, Zhong, Yang, Zhao, Zhou, Li, Ding, et~al.]{bai2025embodied}
Shuanghao Bai, Wenxuan Song, Jiayi Chen, Yuheng Ji, Zhide Zhong, Jin Yang, Han Zhao, Wanqi Zhou, Zhe Li, Pengxiang Ding, et~al.
\newblock Embodied robot manipulation in the era of foundation models: Planning and learning perspectives.
\newblock \emph{arXiv preprint arXiv:2512.22983}, 2025{\natexlab{c}}.

\bibitem[Bai et~al.(2025{\natexlab{d}})Bai, Song, Chen, Ji, Zhong, Yang, Zhao, Zhou, Zhao, Li, et~al.]{bai2025towards}
Shuanghao Bai, Wenxuan Song, Jiayi Chen, Yuheng Ji, Zhide Zhong, Jin Yang, Han Zhao, Wanqi Zhou, Wei Zhao, Zhe Li, et~al.
\newblock Towards a unified understanding of robot manipulation: A comprehensive survey.
\newblock \emph{arXiv preprint arXiv:2510.10903}, 2025{\natexlab{d}}.

\bibitem[Bu et~al.(2025)Bu, Cai, Chen, Cui, Ding, Feng, Gao, He, Huang, Jiang, et~al.]{bu2025agibot}
Qingwen Bu, Jisong Cai, Li~Chen, Xiuqi Cui, Yan Ding, Siyuan Feng, Shenyuan Gao, Xindong He, Xu~Huang, Shu Jiang, et~al.
\newblock Agibot world colosseo: A large-scale manipulation platform for scalable and intelligent embodied systems.
\newblock \emph{arXiv preprint arXiv:2503.06669}, 2025.

\bibitem[Chen et~al.(2024{\natexlab{a}})Chen, Xu, Kirmani, Ichter, Sadigh, Guibas, and Xia]{spatialvlm}
Boyuan Chen, Zhuo Xu, Sean Kirmani, Brain Ichter, Dorsa Sadigh, Leonidas Guibas, and Fei Xia.
\newblock Spatialvlm: Endowing vision-language models with spatial reasoning capabilities.
\newblock In \emph{Proceedings of the IEEE/CVF Conference on Computer Vision and Pattern Recognition}, pages 14455--14465, 2024{\natexlab{a}}.

\bibitem[Chen et~al.(2025{\natexlab{a}})Chen, Lou, Cao, Guo, Fan, Wu, Yang, Ma, and Ye]{chen2025sd}
Pingyi Chen, Yujing Lou, Shen Cao, Jinhui Guo, Lubin Fan, Yue Wu, Lin Yang, Lizhuang Ma, and Jieping Ye.
\newblock Sd-vlm: Spatial measuring and understanding with depth-encoded vision-language models.
\newblock \emph{arXiv preprint arXiv:2509.17664}, 2025{\natexlab{a}}.

\bibitem[Chen et~al.(2025{\natexlab{b}})Chen, Yu, Schwager, Abbeel, Shentu, and Wu]{chen2025sarm}
Qianzhong Chen, Justin Yu, Mac Schwager, Pieter Abbeel, Fred Shentu, and Philipp Wu.
\newblock Sarm: Stage-aware reward modeling for long horizon robot manipulation.
\newblock \emph{arXiv preprint arXiv:2509.25358}, 2025{\natexlab{b}}.

\bibitem[Chen et~al.(2025{\natexlab{c}})Chen, Liu, Qian, Jiang, Li, Zhang, Liu, Gu, Hou, Wang, et~al.]{chen2025ac}
Sixiang Chen, Jiaming Liu, Siyuan Qian, Han Jiang, Lily Li, Renrui Zhang, Zhuoyang Liu, Chenyang Gu, Chengkai Hou, Pengwei Wang, et~al.
\newblock Ac-dit: Adaptive coordination diffusion transformer for mobile manipulation.
\newblock \emph{arXiv preprint arXiv:2507.01961}, 2025{\natexlab{c}}.

\bibitem[Chen et~al.(2025{\natexlab{d}})Chen, Chen, Chen, Cai, Liu, Li, Liang, Lin, Ge, Gu, et~al.]{chen2025robotwin}
Tianxing Chen, Zanxin Chen, Baijun Chen, Zijian Cai, Yibin Liu, Zixuan Li, Qiwei Liang, Xianliang Lin, Yiheng Ge, Zhenyu Gu, et~al.
\newblock Robotwin 2.0: A scalable data generator and benchmark with strong domain randomization for robust bimanual robotic manipulation.
\newblock \emph{arXiv preprint arXiv:2506.18088}, 2025{\natexlab{d}}.

\bibitem[Chen et~al.(2024{\natexlab{b}})Chen, Ge, Ge, Ding, Li, Wang, Xu, Shan, and Liu]{chen2024egoplanbenchbenchmarkingmultimodallarge}
Yi~Chen, Yuying Ge, Yixiao Ge, Mingyu Ding, Bohao Li, Rui Wang, Ruifeng Xu, Ying Shan, and Xihui Liu.
\newblock Egoplan-bench: Benchmarking multimodal large language models for human-level planning, 2024{\natexlab{b}}.
\newblock URL \url{https://arxiv.org/abs/2312.06722}.

\bibitem[contributors(2024)]{contributors2024agibotworldrepo}
AgiBot World~Colosseum contributors.
\newblock Agibot world colosseum.
\newblock \url{https://github.com/OpenDriveLab/AgiBot-World}, 2024.

\bibitem[Contributors(2024)]{flagscale}
FlagScale Contributors.
\newblock Flagscale: A unified meta-framework enabling adaptive heterogeneous computing for the llm ecosystem.
\newblock \url{https://github.com/FlagOpen/FlagScale}, 2024.
\newblock Accessed: 2025-06-26.

\bibitem[Dai et~al.(2017)Dai, Chang, Savva, Halber, Funkhouser, and Nie{\ss}ner]{dai2017scannet}
Angela Dai, Angel~X Chang, Manolis Savva, Maciej Halber, Thomas Funkhouser, and Matthias Nie{\ss}ner.
\newblock Scannet: Richly-annotated 3d reconstructions of indoor scenes.
\newblock In \emph{CVPR}, 2017.

\bibitem[Deitke et~al.(2024)Deitke, Clark, Lee, Tripathi, Yang, Park, Salehi, Muennighoff, Lo, Soldaini, et~al.]{deitke2024molmo}
Matt Deitke, Christopher Clark, Sangho Lee, Rohun Tripathi, Yue Yang, Jae~Sung Park, Mohammadreza Salehi, Niklas Muennighoff, Kyle Lo, Luca Soldaini, et~al.
\newblock Molmo and pixmo: Open weights and open data for state-of-the-art multimodal models.
\newblock \emph{arXiv preprint arXiv:2409.17146}, 2024.

\bibitem[Deitke et~al.(2025)Deitke, Clark, Lee, Tripathi, Yang, Park, Salehi, Muennighoff, Lo, Soldaini, et~al.]{deitke2025molmo}
Matt Deitke, Christopher Clark, Sangho Lee, Rohun Tripathi, Yue Yang, Jae~Sung Park, Mohammadreza Salehi, Niklas Muennighoff, Kyle Lo, Luca Soldaini, et~al.
\newblock Molmo and pixmo: Open weights and open data for state-of-the-art vision-language models.
\newblock In \emph{Proceedings of the Computer Vision and Pattern Recognition Conference}, pages 91--104, 2025.

\bibitem[Du et~al.(2024)Du, Wu, Li, Huang, and Wei]{du2024embspatial}
Mengfei Du, Binhao Wu, Zejun Li, Xuan-Jing Huang, and Zhongyu Wei.
\newblock Embspatial-bench: Benchmarking spatial understanding for embodied tasks with large vision-language models.
\newblock In \emph{ACL}, 2024.

\bibitem[FlagOpen(2025)]{FlagOpen_RoboBrainX0}
FlagOpen.
\newblock Robobrain-x0.
\newblock \url{https://github.com/FlagOpen/RoboBrain-X0}, 2025.
\newblock GitHub repository, accessed 2025-11-08.

\bibitem[Google(2025)]{gemini3pro}
Google.
\newblock Gemini 3 pro: the frontier of vision ai.
\newblock \url{https://blog.google/innovation-and-ai/technology/developers-tools/gemini-3-pro-vision/}, 2025.
\newblock Accessed: 2025-05-06.

\bibitem[Gupta et~al.(2019)Gupta, Dollar, and Girshick]{gupta2019lvis}
Agrim Gupta, Piotr Dollar, and Ross Girshick.
\newblock Lvis: A dataset for large vocabulary instance segmentation.
\newblock In \emph{Proceedings of the IEEE/CVF conference on computer vision and pattern recognition}, pages 5356--5364, 2019.

\bibitem[Hao et~al.(2025)Hao, Zhou, Huang, Hou, Tang, Zhang, Li, Lu, Ren, Meng, et~al.]{hao2025mimo}
Xiaoshuai Hao, Lei Zhou, Zhijian Huang, Zhiwen Hou, Yingbo Tang, Lingfeng Zhang, Guang Li, Zheng Lu, Shuhuai Ren, Xianhui Meng, et~al.
\newblock Mimo-embodied: X-embodied foundation model technical report.
\newblock \emph{arXiv preprint arXiv:2511.16518}, 2025.

\bibitem[He et~al.(2025)He, Liu, shu Zheng, Li, Yao, Qin, Xuan, and Yang]{FlagEvalMM}
Zheqi He, Yesheng Liu, Jing shu Zheng, Xuejing Li, Jin-Ge Yao, Bowen Qin, Richeng Xuan, and Xi~Yang.
\newblock Flagevalmm: A flexible framework for comprehensive multimodal model evaluation.
\newblock 2025.
\newblock URL \url{https://arxiv.org/abs/2506.09081}.

\bibitem[Hoque et~al.(2025)Hoque, Huang, Yoon, Sivapurapu, and Zhang]{hoque2025egodex}
Ryan Hoque, Peide Huang, David~J Yoon, Mouli Sivapurapu, and Jian Zhang.
\newblock Egodex: Learning dexterous manipulation from large-scale egocentric video.
\newblock \emph{arXiv preprint arXiv:2505.11709}, 2025.

\bibitem[Hurst et~al.(2024)Hurst, Lerer, Goucher, Perelman, Ramesh, Clark, Ostrow, Welihinda, Hayes, Radford, et~al.]{gpt4o}
Aaron Hurst, Adam Lerer, Adam~P Goucher, Adam Perelman, Aditya Ramesh, Aidan Clark, AJ~Ostrow, Akila Welihinda, Alan Hayes, Alec Radford, et~al.
\newblock Gpt-4o system card.
\newblock \emph{arXiv preprint arXiv:2410.21276}, 2024.

\bibitem[Intelligence et~al.(2025)Intelligence, Black, Brown, Darpinian, Dhabalia, Driess, Esmail, Equi, Finn, Fusai, et~al.]{pi05}
Physical Intelligence, Kevin Black, Noah Brown, James Darpinian, Karan Dhabalia, Danny Driess, Adnan Esmail, Michael Equi, Chelsea Finn, Niccolo Fusai, et~al.
\newblock pi0.5: a vision-language-action model with open-world generalization.
\newblock \emph{arXiv preprint arXiv:2504.16054}, 2025.

\bibitem[Ji et~al.(2025{\natexlab{a}})Ji, Tan, Shi, Hao, Zhang, Zhang, Wang, Zhao, Mu, An, et~al.]{ji2025robobrain}
Yuheng Ji, Huajie Tan, Jiayu Shi, Xiaoshuai Hao, Yuan Zhang, Hengyuan Zhang, Pengwei Wang, Mengdi Zhao, Yao Mu, Pengju An, et~al.
\newblock Robobrain: A unified brain model for robotic manipulation from abstract to concrete.
\newblock In \emph{Proceedings of the Computer Vision and Pattern Recognition Conference}, pages 1724--1734, 2025{\natexlab{a}}.

\bibitem[Ji et~al.(2025{\natexlab{b}})Ji, Wang, Liu, Hao, Liu, Zhao, Lyu, and Zheng]{ji2025visualtrans}
Yuheng Ji, Yipu Wang, Yuyang Liu, Xiaoshuai Hao, Yue Liu, Yuting Zhao, Huaihai Lyu, and Xiaolong Zheng.
\newblock Visualtrans: A benchmark for real-world visual transformation reasoning.
\newblock \emph{arXiv preprint arXiv:2508.04043}, 2025{\natexlab{b}}.

\bibitem[Jiang et~al.(2025)Jiang, Yuan, Liu, Lu, Cui, Liu, Cheng, Gao, Xu, and Zhao]{jiang2025galaxea}
Tao Jiang, Tianyuan Yuan, Yicheng Liu, Chenhao Lu, Jianning Cui, Xiao Liu, Shuiqi Cheng, Jiyang Gao, Huazhe Xu, and Hang Zhao.
\newblock Galaxea open-world dataset and g0 dual-system vla model.
\newblock \emph{arXiv preprint arXiv:2509.00576}, 2025.

\bibitem[Khazatsky et~al.(2024)Khazatsky, Pertsch, Nair, Balakrishna, Dasari, Karamcheti, Nasiriany, Srirama, Chen, Ellis, et~al.]{khazatsky2024droid}
Alexander Khazatsky, Karl Pertsch, Suraj Nair, Ashwin Balakrishna, Sudeep Dasari, Siddharth Karamcheti, Soroush Nasiriany, Mohan~Kumar Srirama, Lawrence~Yunliang Chen, Kirsty Ellis, et~al.
\newblock Droid: A large-scale in-the-wild robot manipulation dataset.
\newblock \emph{arXiv preprint arXiv:2403.12945}, 2024.

\bibitem[Kolve et~al.(2017)Kolve, Mottaghi, Han, VanderBilt, Weihs, Herrasti, Deitke, Ehsani, Gordon, Zhu, et~al.]{kolve2017ai2}
Eric Kolve, Roozbeh Mottaghi, Winson Han, Eli VanderBilt, Luca Weihs, Alvaro Herrasti, Matt Deitke, Kiana Ehsani, Daniel Gordon, Yuke Zhu, et~al.
\newblock Ai2-thor: An interactive 3d environment for visual ai.
\newblock \emph{arXiv preprint arXiv:1712.05474}, 2017.

\bibitem[Kuznetsova et~al.(2020)Kuznetsova, Rom, Alldrin, Uijlings, Krasin, Pont-Tuset, Kamali, Popov, Malloci, Kolesnikov, et~al.]{kuznetsova2020open}
Alina Kuznetsova, Hassan Rom, Neil Alldrin, Jasper Uijlings, Ivan Krasin, Jordi Pont-Tuset, Shahab Kamali, Stefan Popov, Matteo Malloci, Alexander Kolesnikov, et~al.
\newblock The open images dataset v4: Unified image classification, object detection, and visual relationship detection at scale.
\newblock \emph{IJCV}, 2020.

\bibitem[Lazarow et~al.(2024)Lazarow, Griffiths, Kohavi, Crespo, and Dehghan]{lazarow2024cubify}
Justin Lazarow, David Griffiths, Gefen Kohavi, Francisco Crespo, and Afshin Dehghan.
\newblock Cubify anything: Scaling indoor 3d object detection.
\newblock \emph{arXiv preprint arXiv:2412.04458}, 2024.

\bibitem[Li et~al.(2023)Li, Mai, Liang, and Zaharia]{megatron-energon}
Xuechen Li, Yifan Mai, Percy Liang, and Matei Zaharia.
\newblock Energon: Scaling megatron-lm training with data and expert parallelism, 2023.
\newblock URL \url{https://github.com/HazyResearch/megatron-energon}.

\bibitem[Li et~al.(2024)Li, Yuan, He, Qiu, Zhu, Gu, Shen, Dong, Dong, and Yang]{li2024lamp}
Zhe Li, Weihao Yuan, Yisheng He, Lingteng Qiu, Shenhao Zhu, Xiaodong Gu, Weichao Shen, Yuan Dong, Zilong Dong, and Laurence~T Yang.
\newblock Lamp: Language-motion pretraining for motion generation, retrieval, and captioning.
\newblock \emph{arXiv preprint arXiv:2410.07093}, 2024.

\bibitem[Li et~al.(2025{\natexlab{a}})Li, Chi, Wei, Zhu, Huang, Sun, Peng, Wang, Wang, Liu, et~al.]{li2025robomirror}
Zhe Li, Cheng Chi, Yangyang Wei, Boan Zhu, Tao Huang, Zhenguo Sun, Yibo Peng, Pengwei Wang, Zhongyuan Wang, Fangzhou Liu, et~al.
\newblock Robomirror: Understand before you imitate for video to humanoid locomotion.
\newblock \emph{arXiv preprint arXiv:2512.23649}, 2025{\natexlab{a}}.

\bibitem[Li et~al.(2025{\natexlab{b}})Li, Chi, Wei, Zhu, Huang, Sun, Peng, Wang, Wang, Liu, et~al.]{li2025you}
Zhe Li, Cheng Chi, Yangyang Wei, Boan Zhu, Tao Huang, Zhenguo Sun, Yibo Peng, Pengwei Wang, Zhongyuan Wang, Fangzhou Liu, et~al.
\newblock Do you have freestyle? expressive humanoid locomotion via audio control.
\newblock \emph{arXiv preprint arXiv:2512.23650}, 2025{\natexlab{b}}.

\bibitem[Li et~al.(2025{\natexlab{c}})Li, Chi, Wei, Zhu, Peng, Huang, Wang, Wang, Zhang, and Xu]{li2025language}
Zhe Li, Cheng Chi, Yangyang Wei, Boan Zhu, Yibo Peng, Tao Huang, Pengwei Wang, Zhongyuan Wang, Shanghang Zhang, and Chang Xu.
\newblock From language to locomotion: Retargeting-free humanoid control via motion latent guidance.
\newblock \emph{arXiv preprint arXiv:2510.14952}, 2025{\natexlab{c}}.

\bibitem[Liao et~al.(2024)Liao, Mahmood, Fidler, and Acuna]{liao2024reasoning}
Yuan-Hong Liao, Rafid Mahmood, Sanja Fidler, and David Acuna.
\newblock Reasoning paths with reference objects elicit quantitative spatial reasoning in large vision-language models.
\newblock \emph{arXiv preprint arXiv:2409.09788}, 2024.

\bibitem[Liu et~al.(2024{\natexlab{a}})Liu, Feng, Xue, Wang, Wu, Lu, Zhao, Deng, Zhang, Ruan, et~al.]{deepseekv3}
Aixin Liu, Bei Feng, Bing Xue, Bingxuan Wang, Bochao Wu, Chengda Lu, Chenggang Zhao, Chengqi Deng, Chenyu Zhang, Chong Ruan, et~al.
\newblock Deepseek-v3 technical report.
\newblock \emph{arXiv preprint arXiv:2412.19437}, 2024{\natexlab{a}}.

\bibitem[Liu et~al.(2023)Liu, Zhu, Gao, Feng, Liu, Zhu, and Stone]{liu2023libero}
Bo~Liu, Yifeng Zhu, Chongkai Gao, Yihao Feng, Qiang Liu, Yuke Zhu, and Peter Stone.
\newblock Libero: Benchmarking knowledge transfer for lifelong robot learning.
\newblock \emph{Advances in Neural Information Processing Systems}, 36:\penalty0 44776--44791, 2023.

\bibitem[Liu et~al.(2024{\natexlab{b}})Liu, Li, Li, and Lee]{liu2024improved}
Haotian Liu, Chunyuan Li, Yuheng Li, and Yong~Jae Lee.
\newblock Improved baselines with visual instruction tuning.
\newblock In \emph{Proceedings of the IEEE/CVF Conference on Computer Vision and Pattern Recognition}, pages 26296--26306, 2024{\natexlab{b}}.

\bibitem[Liu et~al.(2024{\natexlab{c}})Liu, Zeng, Ren, Li, Zhang, Yang, Jiang, Li, Yang, Su, et~al.]{liu2024grounding}
Shilong Liu, Zhaoyang Zeng, Tianhe Ren, Feng Li, Hao Zhang, Jie Yang, Qing Jiang, Chunyuan Li, Jianwei Yang, Hang Su, et~al.
\newblock Grounding dino: Marrying dino with grounded pre-training for open-set object detection.
\newblock In \emph{ECCV}, 2024{\natexlab{c}}.

\bibitem[Lyu et~al.(2024)Lyu, Wang, Lin, Yang, Mao, Chen, Xu, Huang, Zhu, Lin, and Pang]{mmscan}
Ruiyuan Lyu, Tai Wang, Jingli Lin, Shuai Yang, Xiaohan Mao, Yilun Chen, Runsen Xu, Haifeng Huang, Chenming Zhu, Dahua Lin, and Jiangmiao Pang.
\newblock Mmscan: A multi-modal 3d scene dataset with hierarchical grounded language annotations.
\newblock \emph{arXiv preprint arXiv:2406.09401}, 2024.

\bibitem[Ma et~al.(2023{\natexlab{a}})Ma, Yong, Zheng, Li, Liang, Zhu, and Huang]{sqa3d}
Xiaojian Ma, Silong Yong, Zilong Zheng, Qing Li, Yitao Liang, Song-Chun Zhu, and Siyuan Huang.
\newblock Sqa3d: Situated question answering in 3d scenes.
\newblock In \emph{ICLR}, 2023{\natexlab{a}}.
\newblock URL \url{https://openreview.net/forum?id=IDJx97BC38}.

\bibitem[Ma et~al.(2023{\natexlab{b}})Ma, Kumar, Zhang, Bastani, and Jayaraman]{ma2023liv}
Yecheng~Jason Ma, Vikash Kumar, Amy Zhang, Osbert Bastani, and Dinesh Jayaraman.
\newblock Liv: Language-image representations and rewards for robotic control.
\newblock In \emph{International Conference on Machine Learning}, pages 23301--23320. PMLR, 2023{\natexlab{b}}.

\bibitem[Ma et~al.(2023{\natexlab{c}})Ma, Liang, Wang, Huang, Bastani, Jayaraman, Zhu, Fan, and Anandkumar]{ma2023eureka}
Yecheng~Jason Ma, William Liang, Guanzhi Wang, De-An Huang, Osbert Bastani, Dinesh Jayaraman, Yuke Zhu, Linxi Fan, and Anima Anandkumar.
\newblock Eureka: Human-level reward design via coding large language models.
\newblock \emph{arXiv preprint arXiv:2310.12931}, 2023{\natexlab{c}}.

\bibitem[Ma et~al.(2024)Ma, Hejna, Fu, Shah, Liang, Xu, Kirmani, Xu, Driess, Xiao, et~al.]{ma2024gvl}
Yecheng~Jason Ma, Joey Hejna, Chuyuan Fu, Dhruv Shah, Jacky Liang, Zhuo Xu, Sean Kirmani, Peng Xu, Danny Driess, Ted Xiao, et~al.
\newblock Vision language models are in-context value learners.
\newblock In \emph{The Thirteenth International Conference on Learning Representations}, 2024.

\bibitem[Nasiriany et~al.(2024)Nasiriany, Maddukuri, Zhang, Parikh, Lo, Joshi, Mandlekar, and Zhu]{nasiriany2024robocasa}
Soroush Nasiriany, Abhiram Maddukuri, Lance Zhang, Adeet Parikh, Aaron Lo, Abhishek Joshi, Ajay Mandlekar, and Yuke Zhu.
\newblock Robocasa: Large-scale simulation of everyday tasks for generalist robots.
\newblock \emph{arXiv preprint arXiv:2406.02523}, 2024.

\bibitem[NVIDIA(2021)]{megatron}
NVIDIA.
\newblock Megatron-lm: Training multi-billion parameter language models using model parallelism, 2021.
\newblock URL \url{https://github.com/NVIDIA/Megatron-LM}.

\bibitem[OpenAI(2025)]{gpt5.2}
OpenAI.
\newblock Update to gpt-5 system card: Gpt-5.2.
\newblock \url{https://openai.com/zh-Hans-CN/index/gpt-5-system-card-update-gpt-5-2/}, 2025.
\newblock Accessed: 2025-05-06.

\bibitem[Ouyang(2025)]{ouyang2025spatial}
Kun Ouyang.
\newblock Spatial-r1: Enhancing mllms in video spatial reasoning.
\newblock \emph{arXiv preprint arXiv:2504.01805}, 2025.

\bibitem[O’Neill et~al.(2024)O’Neill, Rehman, Maddukuri, Gupta, Padalkar, Lee, Pooley, Gupta, Mandlekar, Jain, et~al.]{o2024open}
Abby O’Neill, Abdul Rehman, Abhiram Maddukuri, Abhishek Gupta, Abhishek Padalkar, Abraham Lee, Acorn Pooley, Agrim Gupta, Ajay Mandlekar, Ajinkya Jain, et~al.
\newblock Open x-embodiment: Robotic learning datasets and rt-x models: Open x-embodiment collaboration 0.
\newblock In \emph{2024 IEEE International Conference on Robotics and Automation (ICRA)}, pages 6892--6903. IEEE, 2024.

\bibitem[Piccinelli et~al.(2025)Piccinelli, Sakaridis, Yang, Segu, Li, Abbeloos, and Van~Gool]{piccinelli2025unidepthv2}
Luigi Piccinelli, Christos Sakaridis, Yung-Hsu Yang, Mattia Segu, Siyuan Li, Wim Abbeloos, and Luc Van~Gool.
\newblock Unidepthv2: Universal monocular metric depth estimation made simpler.
\newblock \emph{arXiv}, 2025.

\bibitem[{PyTorch Developers}(2023)]{pytorch-memory}
{PyTorch Developers}.
\newblock Cuda memory management, 2023.
\newblock URL \url{https://pytorch.org/docs/stable/notes/cuda.html#cuda-memory-management}.

\bibitem[{Qwen Team}(2025)]{qwen2.5-vl}
{Qwen Team}.
\newblock Qwen2.5-vl: Multimodal llms from alibaba, 2025.
\newblock URL \url{https://github.com/QwenLM/Qwen2.5-VL}.

\bibitem[Ramanathan et~al.(2023)Ramanathan, Kalia, Petrovic, Wen, Zheng, Guo, Wang, Marquez, Kovvuri, Kadian, et~al.]{ramanathan2023paco}
Vignesh Ramanathan, Anmol Kalia, Vladan Petrovic, Yi~Wen, Baixue Zheng, Baishan Guo, Rui Wang, Aaron Marquez, Rama Kovvuri, Abhishek Kadian, et~al.
\newblock Paco: Parts and attributes of common objects.
\newblock In \emph{Proceedings of the IEEE/CVF Conference on Computer Vision and Pattern Recognition}, pages 7141--7151, 2023.

\bibitem[Ravi et~al.(2025)Ravi, Gabeur, Hu, Hu, Ryali, Ma, Khedr, R{\"a}dle, Rolland, Gustafson, et~al.]{ravi2024sam}
Nikhila Ravi, Valentin Gabeur, Yuan-Ting Hu, Ronghang Hu, Chaitanya Ryali, Tengyu Ma, Haitham Khedr, Roman R{\"a}dle, Chloe Rolland, Laura Gustafson, et~al.
\newblock Sam 2: Segment anything in images and videos.
\newblock \emph{ICLR}, 2025.

\bibitem[Sermanet et~al.(2024)Sermanet, Ding, Zhao, Xia, Dwibedi, Gopalakrishnan, Chan, Dulac-Arnold, Maddineni, Joshi, et~al.]{sermanet2024robovqa}
Pierre Sermanet, Tianli Ding, Jeffrey Zhao, Fei Xia, Debidatta Dwibedi, Keerthana Gopalakrishnan, Christine Chan, Gabriel Dulac-Arnold, Sharath Maddineni, Nikhil~J Joshi, et~al.
\newblock Robovqa: Multimodal long-horizon reasoning for robotics.
\newblock In \emph{2024 IEEE International Conference on Robotics and Automation (ICRA)}, pages 645--652. IEEE, 2024.

\bibitem[Song et~al.(2025)Song, Blukis, Tremblay, Tyree, Su, and Birchfield]{song2025robospatial}
Chan~Hee Song, Valts Blukis, Jonathan Tremblay, Stephen Tyree, Yu~Su, and Stan Birchfield.
\newblock Robospatial: Teaching spatial understanding to 2d and 3d vision-language models for robotics.
\newblock In \emph{Proceedings of the Computer Vision and Pattern Recognition Conference}, pages 15768--15780, 2025.

\bibitem[Tan et~al.(2025{\natexlab{a}})Tan, Chen, Xu, Wang, Ji, Chi, Lyu, Zhao, Chen, Co, et~al.]{tan2025robo}
Huajie Tan, Sixiang Chen, Yijie Xu, Zixiao Wang, Yuheng Ji, Cheng Chi, Yaoxu Lyu, Zhongxia Zhao, Xiansheng Chen, Peterson Co, et~al.
\newblock Robo-dopamine: General process reward modeling for high-precision robotic manipulation.
\newblock \emph{arXiv preprint arXiv:2512.23703}, 2025{\natexlab{a}}.

\bibitem[Tan et~al.(2025{\natexlab{b}})Tan, Chi, Chen, Ji, Zhao, Hao, Lyu, Cao, Zhao, Lyu, et~al.]{tan2025roboosnext}
Huajie Tan, Cheng Chi, Xiansheng Chen, Yuheng Ji, Zhongxia Zhao, Xiaoshuai Hao, Yaoxu Lyu, Mingyu Cao, Junkai Zhao, Huaihai Lyu, et~al.
\newblock Roboos-next: A unified memory-based framework for lifelong, scalable, and robust multi-robot collaboration.
\newblock \emph{arXiv preprint arXiv:2510.26536}, 2025{\natexlab{b}}.

\bibitem[Tan et~al.(2025{\natexlab{c}})Tan, Hao, Lin, Wang, Lyu, Cao, Wang, and Zhang]{tan2025roboos}
Huajie Tan, Xiaoshuai Hao, Minglan Lin, Pengwei Wang, Yaoxu Lyu, Mingyu Cao, Zhongyuan Wang, and Shanghang Zhang.
\newblock Roboos: A hierarchical embodied framework for cross-embodiment and multi-agent collaboration.
\newblock \emph{arXiv preprint arXiv:2505.03673}, 2025{\natexlab{c}}.

\bibitem[Tan et~al.(2025{\natexlab{d}})Tan, Ji, Hao, Chen, Wang, Wang, and Zhang]{tan2025reason}
Huajie Tan, Yuheng Ji, Xiaoshuai Hao, Xiansheng Chen, Pengwei Wang, Zhongyuan Wang, and Shanghang Zhang.
\newblock Reason-rft: Reinforcement fine-tuning for visual reasoning of vision language models.
\newblock In \emph{The Thirty-ninth Annual Conference on Neural Information Processing Systems}, 2025{\natexlab{d}}.

\bibitem[Tan et~al.(2026)Tan, Co, Xu, Rong, Ji, Chi, Chen, Zhang, Zhao, Wang, et~al.]{tan2026action}
Huajie Tan, Peterson Co, Yijie Xu, Shanyu Rong, Yuheng Ji, Cheng Chi, Xiansheng Chen, Qiongyu Zhang, Zhongxia Zhao, Pengwei Wang, et~al.
\newblock Action-sketcher: From reasoning to action via visual sketches for long-horizon robotic manipulation.
\newblock \emph{arXiv preprint arXiv:2601.01618}, 2026.

\bibitem[Team et~al.(2025{\natexlab{a}})Team, Cao, Tan, Ji, Chen, Lin, Li, Cao, Wang, Zhou, et~al.]{team2025robobrain}
BAAI~RoboBrain Team, Mingyu Cao, Huajie Tan, Yuheng Ji, Xiansheng Chen, Minglan Lin, Zhiyu Li, Zhou Cao, Pengwei Wang, Enshen Zhou, et~al.
\newblock Robobrain 2.0 technical report.
\newblock \emph{arXiv preprint arXiv:2507.02029}, 2025{\natexlab{a}}.

\bibitem[Team et~al.(2025{\natexlab{b}})Team, Abeyruwan, Ainslie, Alayrac, Arenas, Armstrong, Balakrishna, Baruch, Bauza, Blokzijl, et~al.]{geminirobotics}
Gemini~Robotics Team, Saminda Abeyruwan, Joshua Ainslie, Jean-Baptiste Alayrac, Montserrat~Gonzalez Arenas, Travis Armstrong, Ashwin Balakrishna, Robert Baruch, Maria Bauza, Michiel Blokzijl, et~al.
\newblock Gemini robotics: Bringing ai into the physical world.
\newblock \emph{arXiv preprint arXiv:2503.20020}, 2025{\natexlab{b}}.

\bibitem[Team(2025)]{qwq32b}
Qwen Team.
\newblock Qwq-32b: Embracing the power of reinforcement learning, March 2025.
\newblock URL \url{https://qwenlm.github.io/blog/qwq-32b/}.

\bibitem[Tong et~al.(2024)Tong, Brown, Wu, Woo, IYER, Akula, Yang, Yang, Middepogu, Wang, et~al.]{tong2024cambrian}
Peter Tong, Ellis Brown, Penghao Wu, Sanghyun Woo, Adithya Jairam~Vedagiri IYER, Sai~Charitha Akula, Shusheng Yang, Jihan Yang, Manoj Middepogu, Ziteng Wang, et~al.
\newblock Cambrian-1: A fully open, vision-centric exploration of multimodal llms.
\newblock \emph{NeurIPS}, 2024.

\bibitem[Wald et~al.(2019)Wald, Avetisyan, Navab, Tombari, and Nie{\ss}ner]{3rscan}
Johanna Wald, Armen Avetisyan, Nassir Navab, Federico Tombari, and Matthias Nie{\ss}ner.
\newblock Rio: 3d object instance re-localization in changing indoor environments.
\newblock In \emph{ICCV}, pages 7658--7667, 2019.

\bibitem[Wang et~al.(2025)Wang, Ji, Liu, Zhou, Yang, Tian, Qin, Liu, Tan, Chi, et~al.]{wang2025towards}
Yipu Wang, Yuheng Ji, Yuyang Liu, Enshen Zhou, Ziqiang Yang, Yuxuan Tian, Ziheng Qin, Yue Liu, Huajie Tan, Cheng Chi, et~al.
\newblock Towards cross-view point correspondence in vision-language models.
\newblock \emph{arXiv preprint arXiv:2512.04686}, 2025.

\bibitem[Yuan et~al.(2024)Yuan, Duan, Blukis, Pumacay, Krishna, Murali, Mousavian, and Fox]{yuan2024robopointvisionlanguagemodelspatial}
Wentao Yuan, Jiafei Duan, Valts Blukis, Wilbert Pumacay, Ranjay Krishna, Adithyavairavan Murali, Arsalan Mousavian, and Dieter Fox.
\newblock Robopoint: A vision-language model for spatial affordance prediction for robotics, 2024.
\newblock URL \url{https://arxiv.org/abs/2406.10721}.

\bibitem[Yuan et~al.(2025)Yuan, Cui, Huang, Chen, Ni, Dong, Li, Zheng, and Hao]{yuan2025embodied}
Yifu Yuan, Haiqin Cui, Yaoting Huang, Yibin Chen, Fei Ni, Zibin Dong, Pengyi Li, Yan Zheng, and Jianye Hao.
\newblock Embodied-r1: Reinforced embodied reasoning for general robotic manipulation.
\newblock \emph{arXiv preprint arXiv:2508.13998}, 2025.

\bibitem[Zhai et~al.(2025)Zhai, Zhang, Zhang, Huang, Zhang, Zhou, Zhang, Liu, Lin, and Pang]{zhai2025vlac}
Shaopeng Zhai, Qi~Zhang, Tianyi Zhang, Fuxian Huang, Haoran Zhang, Ming Zhou, Shengzhe Zhang, Litao Liu, Sixu Lin, and Jiangmiao Pang.
\newblock A vision-language-action-critic model for robotic real-world reinforcement learning.
\newblock \emph{arXiv preprint arXiv:2509.15937}, 2025.

\bibitem[Zhang et~al.(2025{\natexlab{a}})Zhang, Wang, Liu, Huixin, Jiang, Shen, Hou, Zheng, Zhang, Li, et~al.]{zhang2025embodied}
Wenqi Zhang, Mengna Wang, Gangao Liu, Xu~Huixin, Yiwei Jiang, Yongliang Shen, Guiyang Hou, Zhe Zheng, Hang Zhang, Xin Li, et~al.
\newblock Embodied-reasoner: Synergizing visual search, reasoning, and action for embodied interactive tasks.
\newblock \emph{arXiv preprint arXiv:2503.21696}, 2025{\natexlab{a}}.

\bibitem[Zhang et~al.(2025{\natexlab{b}})Zhang, Ni, Chen, Zhang, Rao, Peng, Lu, Hu, Guo, and Hu]{zhang2025beehighqualitycorpusfullstack}
Yi~Zhang, Bolin Ni, Xin-Sheng Chen, Heng-Rui Zhang, Yongming Rao, Houwen Peng, Qinglin Lu, Han Hu, Meng-Hao Guo, and Shi-Min Hu.
\newblock Bee: A high-quality corpus and full-stack suite to unlock advanced fully open mllms, 2025{\natexlab{b}}.
\newblock URL \url{https://arxiv.org/abs/2510.13795}.

\bibitem[Zhang et~al.(2024)Zhang, Huang, Ma, Li, Luo, Xie, Qin, Luo, Li, Liu, et~al.]{zhang2024recognize}
Youcai Zhang, Xinyu Huang, Jinyu Ma, Zhaoyang Li, Zhaochuan Luo, Yanchun Xie, Yuzhuo Qin, Tong Luo, Yaqian Li, Shilong Liu, et~al.
\newblock Recognize anything: A strong image tagging model.
\newblock In \emph{CVPR}, 2024.

\bibitem[Zhou et~al.(2024)Zhou, Su, Chi, Zhang, Wang, Huang, Sheng, and Wang]{zhou2024code}
Enshen Zhou, Qi~Su, Cheng Chi, Zhizheng Zhang, Zhongyuan Wang, Tiejun Huang, Lu~Sheng, and He~Wang.
\newblock Code-as-monitor: Constraint-aware visual programming for reactive and proactive robotic failure detection.
\newblock \emph{arXiv preprint arXiv:2412.04455}, 2024.

\bibitem[Zhou et~al.(2025{\natexlab{a}})Zhou, An, Chi, Han, Rong, Zhang, Wang, Wang, Huang, Sheng, et~al.]{zhou2025roborefer}
Enshen Zhou, Jingkun An, Cheng Chi, Yi~Han, Shanyu Rong, Chi Zhang, Pengwei Wang, Zhongyuan Wang, Tiejun Huang, Lu~Sheng, et~al.
\newblock Roborefer: Towards spatial referring with reasoning in vision-language models for robotics.
\newblock \emph{arXiv preprint arXiv:2506.04308}, 2025{\natexlab{a}}.

\bibitem[Zhou et~al.(2025{\natexlab{b}})Zhou, Chi, Li, An, Zhang, Rong, Han, Ji, Liu, Wang, et~al.]{zhou2025robotracer}
Enshen Zhou, Cheng Chi, Yibo Li, Jingkun An, Jiayuan Zhang, Shanyu Rong, Yi~Han, Yuheng Ji, Mengzhen Liu, Pengwei Wang, et~al.
\newblock Robotracer: Mastering spatial trace with reasoning in vision-language models for robotics.
\newblock \emph{arXiv preprint arXiv:2512.13660}, 2025{\natexlab{b}}.

\bibitem[Zhu et~al.(2023)Zhu, Kumar, Hu, and Liu]{zhu2023tame}
Shengjie Zhu, Abhinav Kumar, Masa Hu, and Xiaoming Liu.
\newblock Tame a wild camera: in-the-wild monocular camera calibration.
\newblock \emph{NIPS}, 2023.

\end{thebibliography}

\clearpage
\section{Contributions and Author List}
\label{sec:contributions}
\setlength{\parskip}{0pt}
\setlength{\itemsep}{0pt}
\setlength{\parsep}{0pt}
\begin{multicols}{2}

\subsubsection*{Core Contributors}
\begin{itemize}
    \vspace{1em}\item Huajie Tan$^{*\dagger}$
    \vspace{1em}\item Enshen Zhou$^*$
    \vspace{1em}\item Zhiyu Li
    \vspace{1em}\item Yijie Xu
    \vspace{1em}\item Yuheng Ji
    \vspace{1em}\item Xiansheng Chen
    \vspace{1em}\item Cheng Chi
    \vspace{1em}\item Pengwei Wang$^\dagger$
    \vspace{1em}\item Huizhu Jia
    \vspace{1em}\item Yulong Ao
    \vspace{1em}\item Yonghua Lin
    \vspace{1em}\item Zhongyuan Wang
    \vspace{1em}\item Tiejun Huang
    \vspace{1em}\item Shanghang Zhang$^{\text{\Letter}}$

\end{itemize}

\vspace{15em}
\subsubsection*{Contributors}

\begin{itemize}
    \vspace{1em}\item Mingyu Cao
    \vspace{1em}\item Sixiang Chen
    \vspace{1em}\item Zhe Li
    \vspace{1em}\item Mengzhen Liu
    \vspace{1em}\item Zixiao Wang
    \vspace{1em}\item Shanyu Rong
    \vspace{1em}\item Yaoxu Lyu
    \vspace{1em}\item Zhongxia Zhao
    \vspace{1em}\item Peterson Co
    \vspace{1em}\item Yibo Li
    \vspace{1em}\item Yi Han
    \vspace{1em}\item Shaoxuan Xie
    \vspace{1em}\item Guocai Yao
    \vspace{1em}\item Songjing Wang
    \vspace{1em}\item Leiduo Zhang
    \vspace{1em}\item Xi Yang
    \vspace{1em}\item Yance Jiao
    \vspace{1em}\item Donghai Shi
    \vspace{1em}\item Kunchang Xie
    \vspace{1em}\item Shaokai Nie
    \vspace{1em}\item Chunlei Men
\end{itemize}

\end{multicols}

\let\thefootnote\relax\footnotetext{$^{*}$ Equal Contribution (Co-first Authors).}
\let\thefootnote\relax\footnotetext{$^{\dagger}$ Project Leaders.}
\let\thefootnote\relax\footnotetext{$^{\text{\Letter}}$ Corresponding Author. Team Email: \url{robobrain@baai.ac.cn}}
\clearpage

\beginappendix

\section{Qualitative examples}
This section provides a comprehensive set of qualitative examples that illustrate the capabilities of RoboBrain 2.5 in various embodied AI tasks. As Capabilities like pointing, affordance, planning, etc. are similar to those shown in RoboBrain 2.0~\cite{team2025robobrain}, the examples in this section \textbf{ONLY} demonstrate the model's proficiency in 3D spatial reasoning, temporal value estimation, showcasing its potential for real-world applications.

\label{sec:app:qualitative}

\subsection{Examples on 3D Spatial Reasoning}
In this section, we provide qualitative visualizations to demonstrate the robustness and precision of RoboBrain 2.5’s 3D spatial reasoning capabilities in real-world manipulation scenarios. We focus on three core aspects: compliance with fine-grained spatial constraints, multi-step compositional reasoning for complex manipulation tasks, and generalization across diverse indoor environments and object categories.

\vspace{0.3em}

\begin{figure*}[h]
    \centering
    \includegraphics[width=0.85\linewidth]{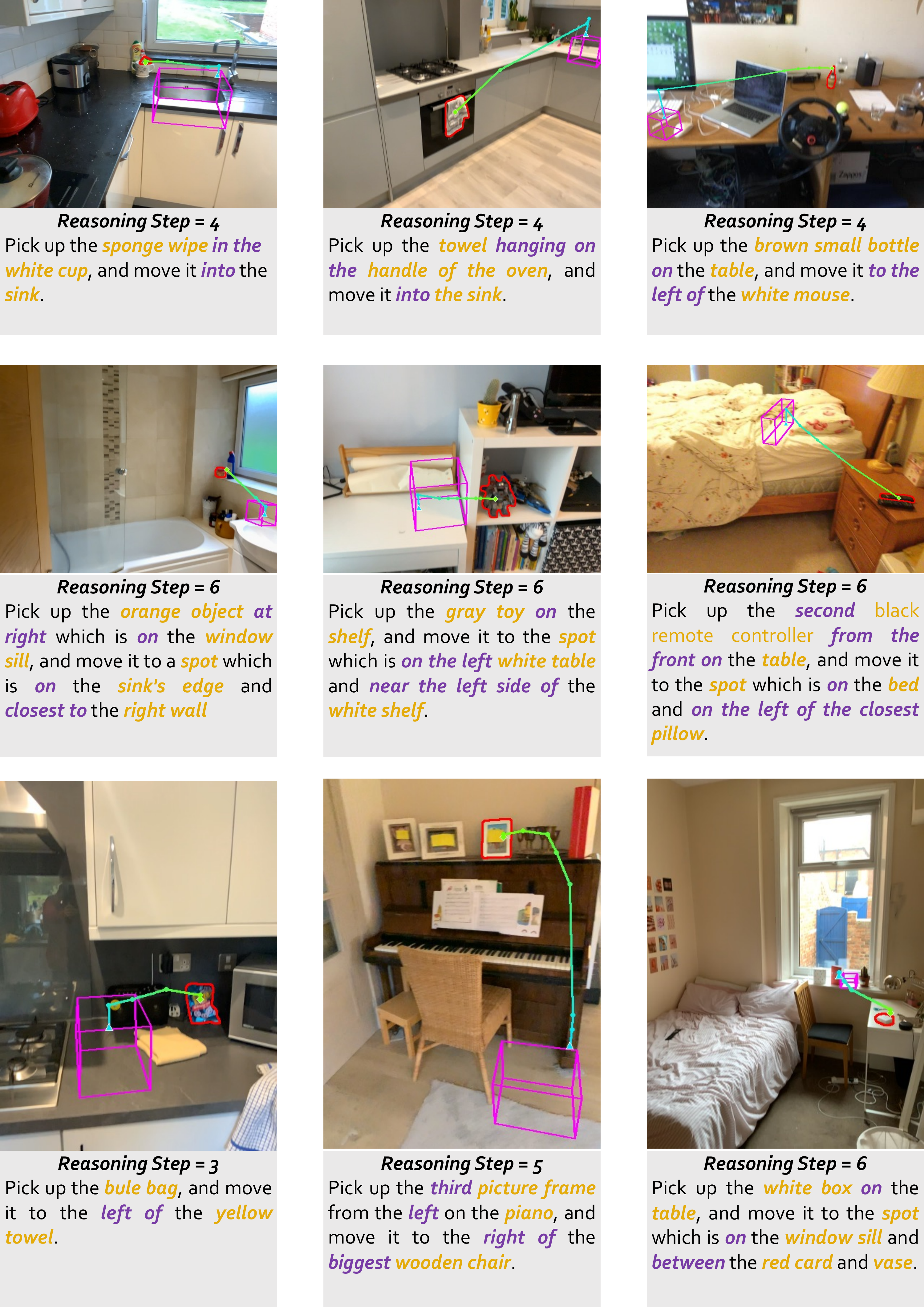}
    \caption{\textbf{Visualization of TraceSpatial-Bench Rollouts and RoboBrain 2.5’s Predicted Traces.} The red mask marks the ground-truth starting point, the purple 3D bounding box represents the ground-truth endpoint, and the 2D projection of RoboBrain 2.5’s predicted 3D spatial trace is displayed.}
    \label{fig:app_3d_trajectory_cases_1}
\end{figure*}

\begin{figure*}[h]
    \centering
    \includegraphics[width=0.85\linewidth]{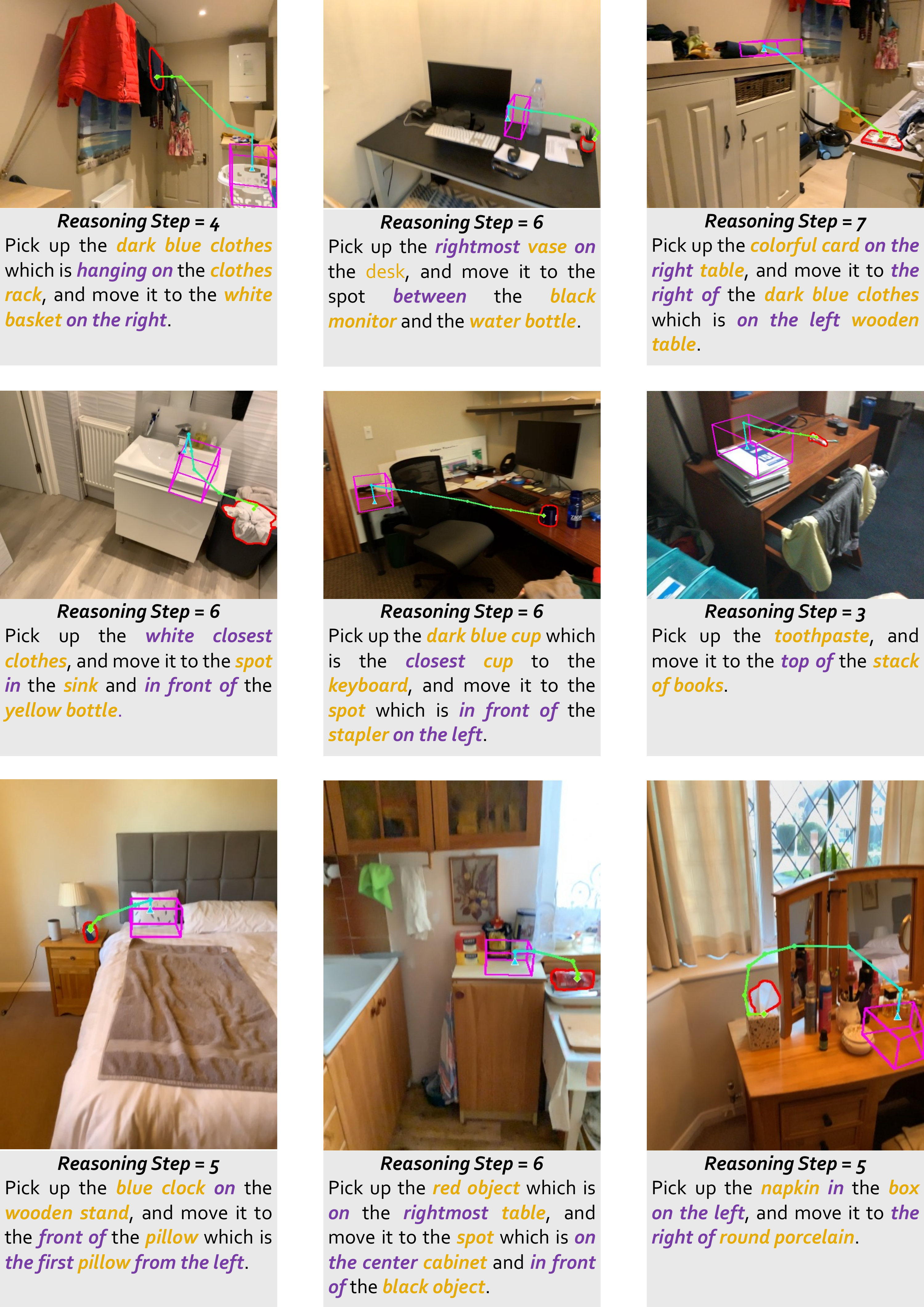}
    \caption{\textbf{Visualization of TraceSpatial-Bench Rollouts and RoboBrain 2.5’s Predicted Traces.} The red mask marks the ground-truth starting point, the purple 3D bounding box represents the ground-truth endpoint, and the 2D projection of RoboBrain 2.5’s predicted 3D spatial trace is displayed.}
    \label{fig:app_3d_trajectory_cases_2}
\end{figure*}

\textbf{Compliance with Fine-Grained Spatial Constraints.}
RoboBrain 2.5 excels at interpreting spatially constrained instructions and generating accurate 3D manipulation traces that adhere to both relative positional requirements and metric constraints. For instructions like “Pick up the third picture frame from the left on the piano, and move it to the right of the biggest wooden chair” in~\Cref{fig:app_3d_trajectory_cases_1} or “Pick up the rightmost vase on the desk, and move it to the spot between the black monitor and the water bottle” in~\Cref{fig:app_3d_trajectory_cases_2}, the model accurately parses ordinal references and spatial relationships to generate collision-free 3D trajectories.

\vspace{0.3em}

\textbf{Multi-Step Compositional Reasoning.}
Complex manipulation tasks often demand multi-step reasoning to decompose high-level goals into executable sub-tasks. RoboBrain 2.5’s 3D spatial tracing capability naturally encodes this compositional logic by generating ordered keypoint sequences. For example, instructions like “Pick up the orange object at right which is on the window sill, and move it to a spot which is on the sink's edge and closest to the right wall” require the model to firstly localize the orange object on the window sill, then estimate the sink’s edge position and its distance to the right wall, and finally generate a smooth 3D trace connecting the two points while avoiding obstacles.

\vspace{0.3em}

\begin{figure*}[h]
    \centering
    \includegraphics[width=0.85\linewidth]{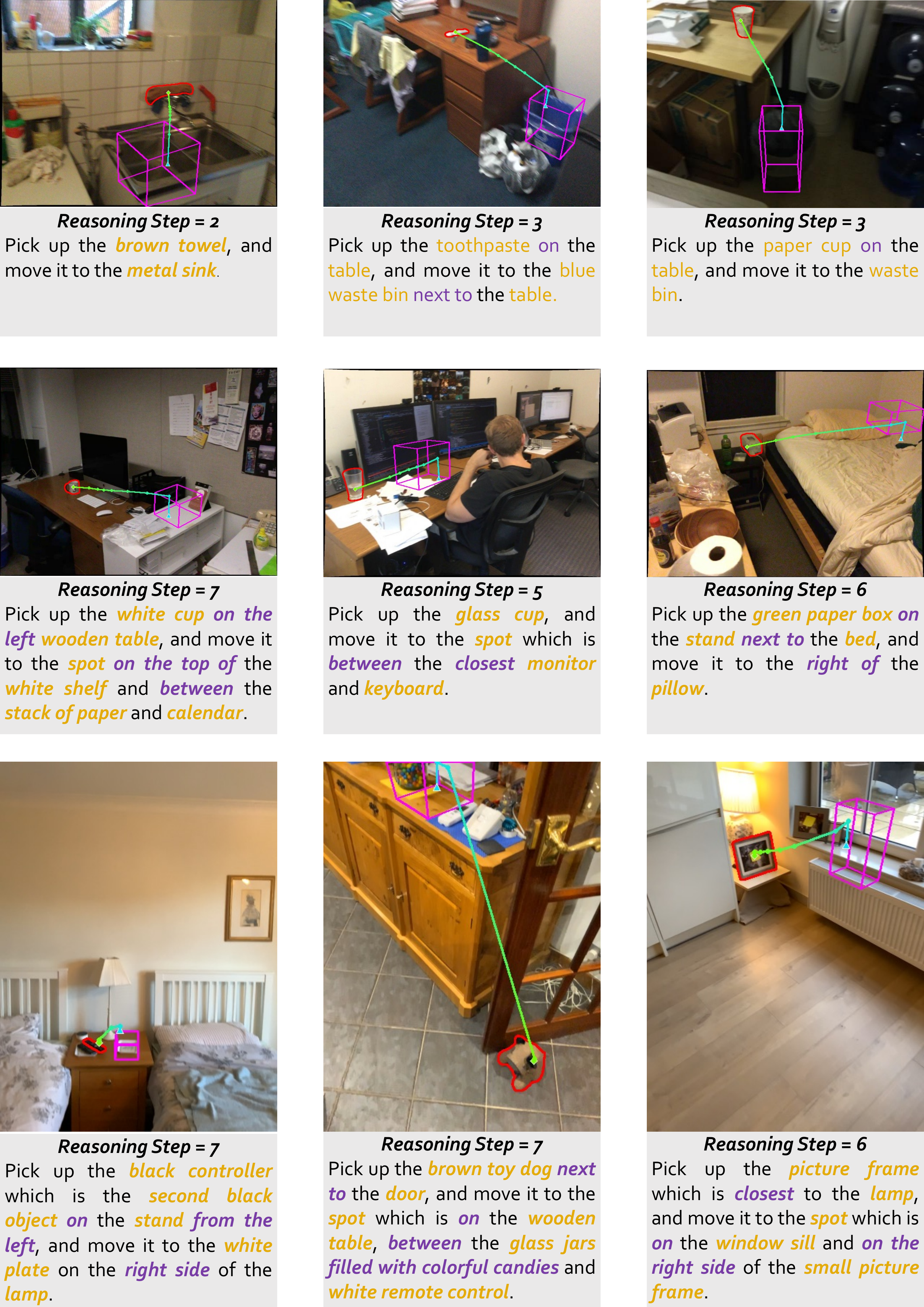}
    \caption{\textbf{Visualization of TraceSpatial-Bench Rollouts and RoboBrain 2.5’s Predicted Traces.} The red mask marks the ground-truth starting point, the purple 3D bounding box represents the ground-truth endpoint, and the 2D projection of RoboBrain 2.5’s predicted 3D spatial trace is displayed.}
    \label{fig:app_3d_trajectory_cases_3}
\end{figure*}

\textbf{Generalization Across Environments and Objects.}
RoboBrain 2.5’s 3D spatial reasoning generalizes to diverse indoor settings and object categories. Tasks span kitchen-centric scenarios in~\Cref{fig:app_3d_trajectory_cases_2}, bedroom settings in~\Cref{fig:app_3d_trajectory_cases_3}, and office/study spaces. The model maintains precision across these environments by leveraging metric-grounded spatial representations independent of scene context. Besides, the model handles objects of varying sizes, shapes, and functionalities, which are from small items to larger objects, and adapts its 3D traces to object-specific affordances.

\vspace{0.3em}

\begin{figure*}[h]
    \centering
    \includegraphics[width=0.85\linewidth]{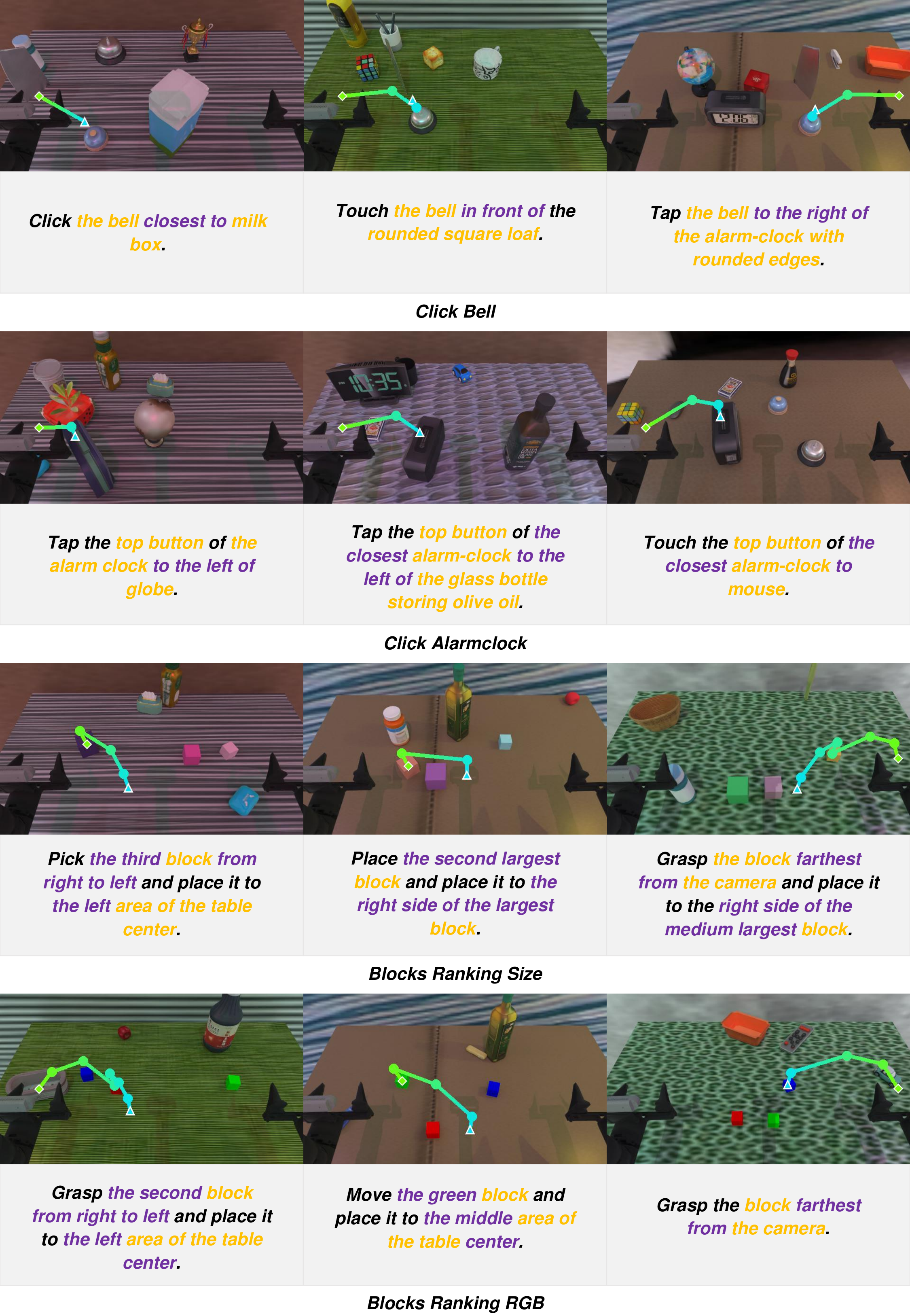}
    \caption{\textbf{Visualization of RoboTwin 2.0 Rollouts and RoboBrain 2.5’s Predicted Traces.} Examples for AgiLex Dual-Arm tasks: Click Bell; Click Alarm clock; Blocks Ranking.}
    \label{fig:app_3d_trajectory_cases_4}
\end{figure*}

\begin{figure*}[h]
    \centering
    \includegraphics[width=0.85\linewidth]{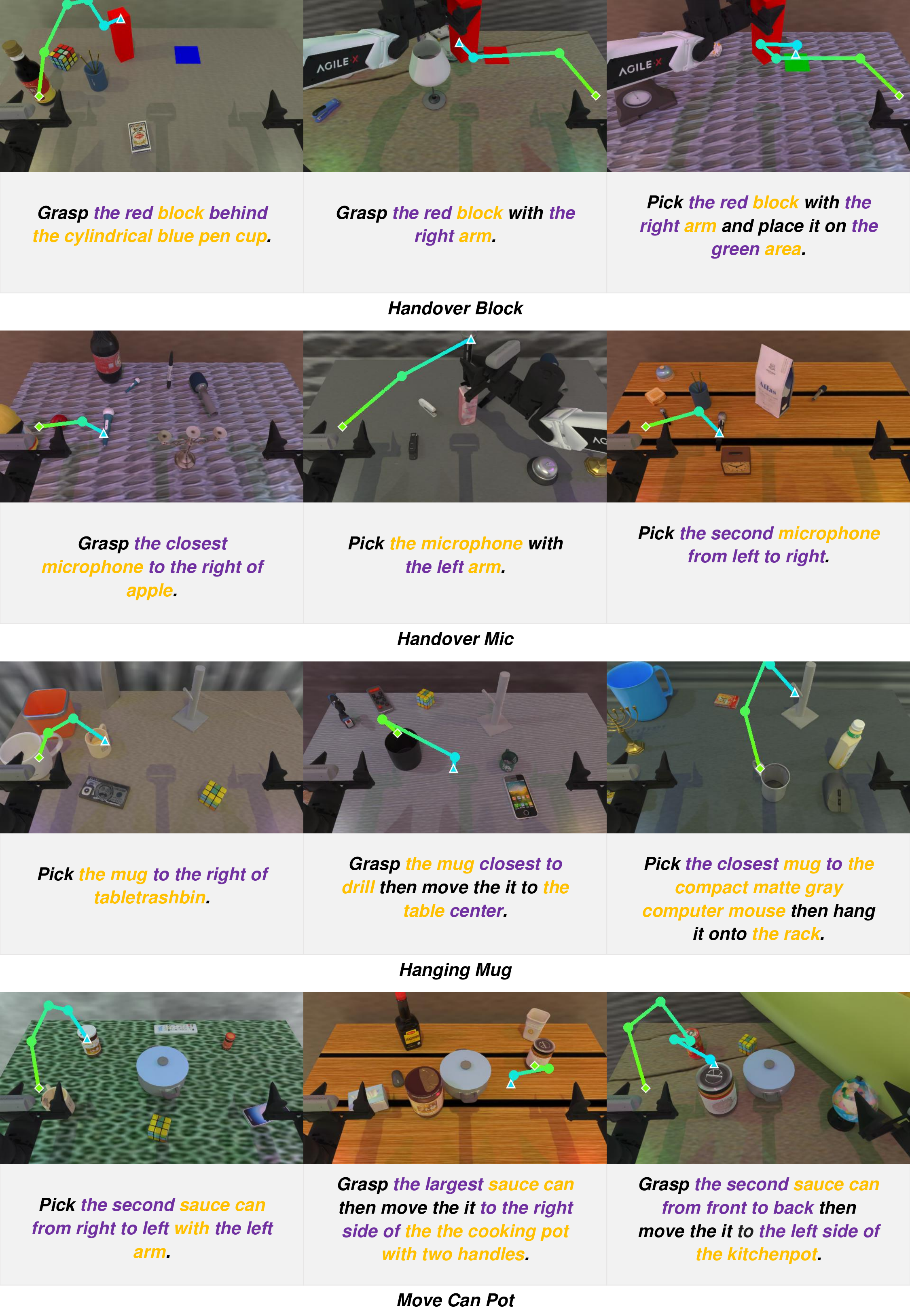}
    \caption{\textbf{Visualization of RoboTwin 2.0 Rollouts and RoboBrain 2.5’s Predicted Traces.} Examples for AgiLex Dual-Arm tasks: Handover Block; Handover Mic; Hanging Mug; Move Can Pot.}
    \label{fig:app_3d_trajectory_cases_5}
\end{figure*}

\begin{figure*}[h]
    \centering
    \includegraphics[width=0.85\linewidth]{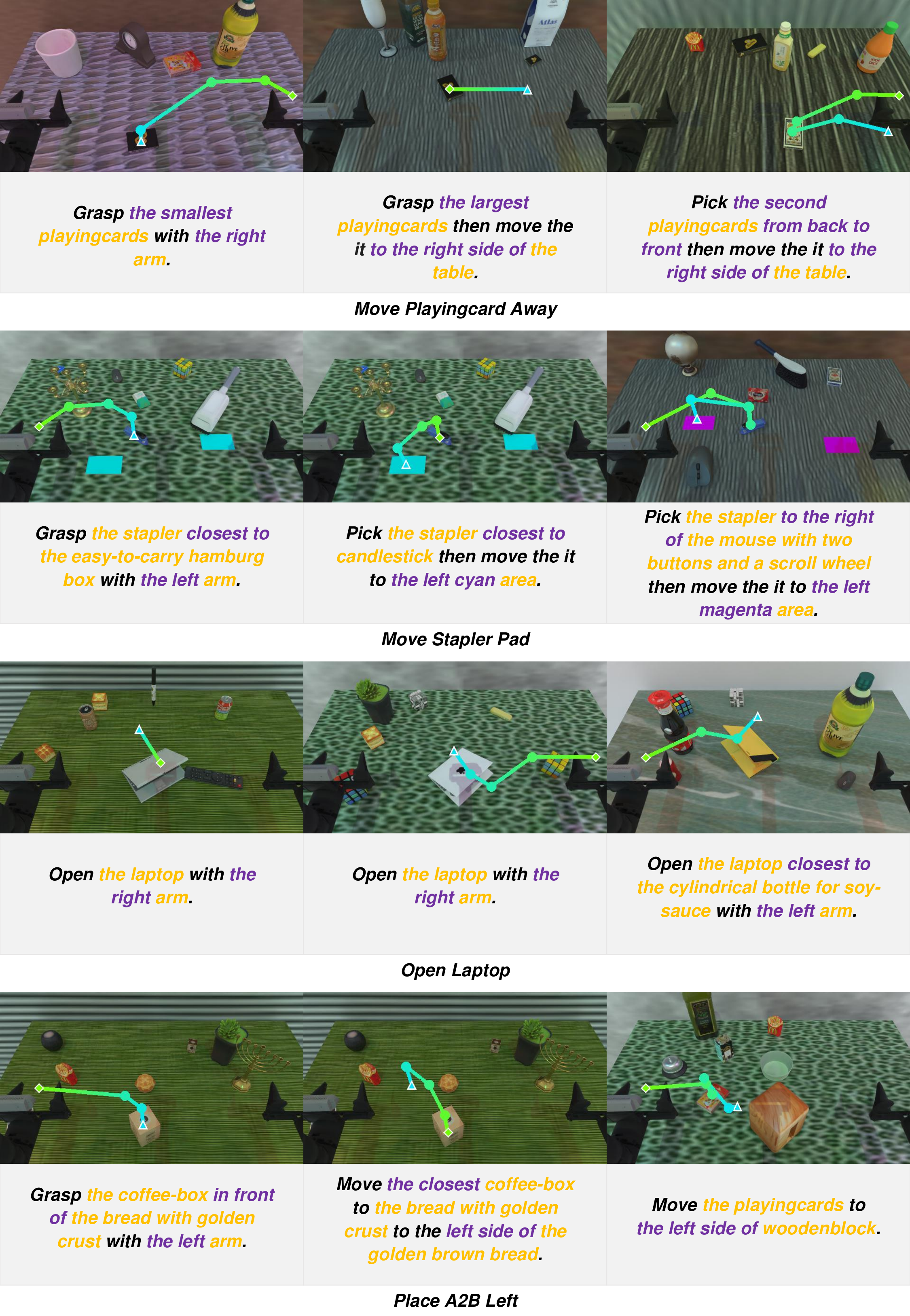}
    \caption{\textbf{Visualization of RoboTwin 2.0 Rollouts and RoboBrain 2.5’s Predicted Traces.} Examples for AgiLex Dual-Arm tasks: Move Playingcard Away; Move Stapler Pad; Open Laptop; Place A2B Left.}
    \label{fig:app_3d_trajectory_cases_6}
\end{figure*}

\begin{figure*}[h]
    \centering
    \includegraphics[width=0.85\linewidth]{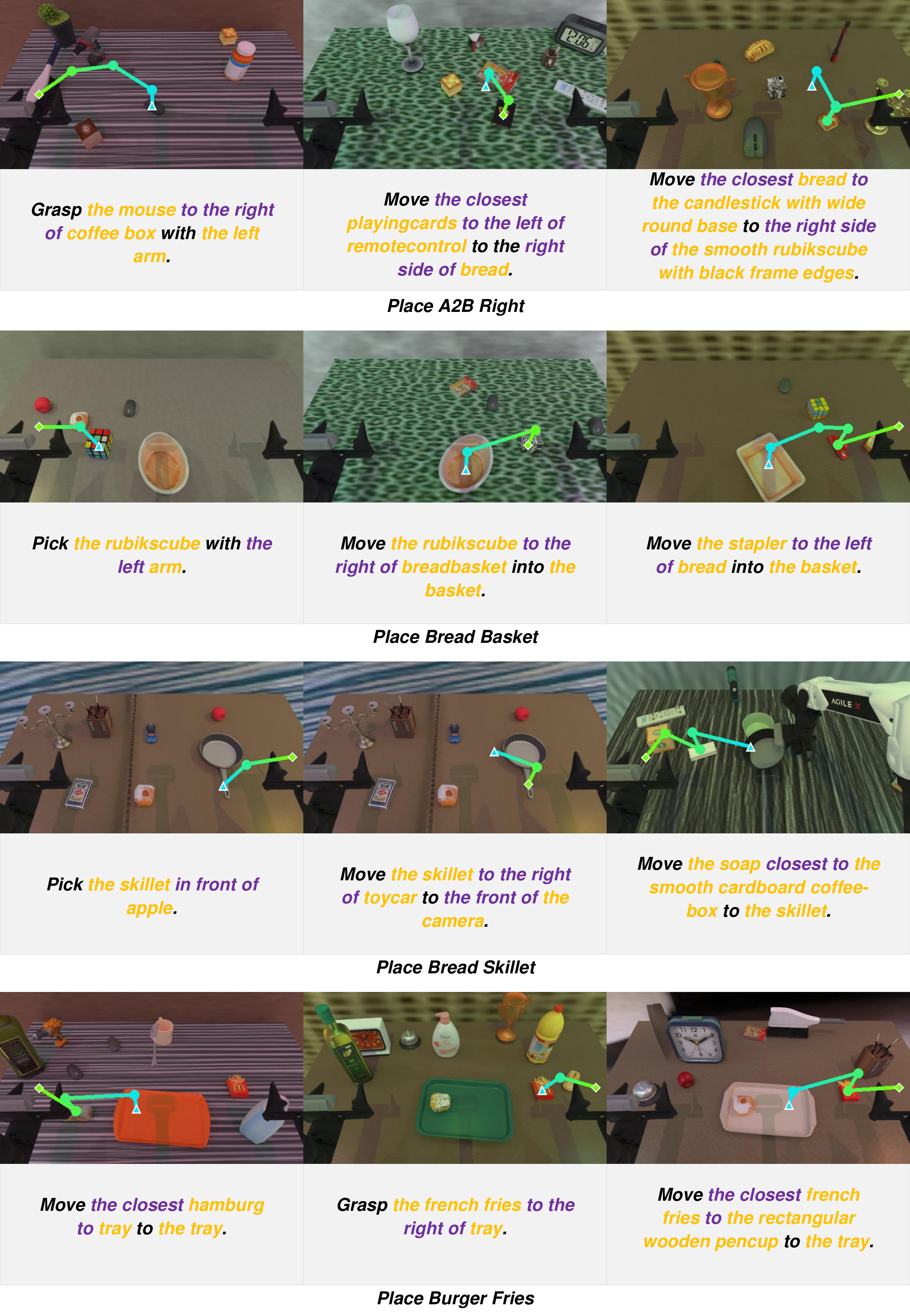}
    \caption{\textbf{Visualization of RoboTwin 2.0 Rollouts and RoboBrain 2.5’s Predicted Traces.} Examples for AgiLex Dual-Arm tasks: Place A2B Right; Place Bread Basket; Place Bread Skillet; Place Burger Fries.}
    \label{fig:app_3d_trajectory_cases_7}
\end{figure*}

\textbf{Application on RoboTwin 2.0}
To further validate the generalization capabilities of RoboBrain 2.5 in simulated collaborative environments, we present qualitative results on the RoboTwin 2.0~\cite{chen2025robotwin}, specifically focusing on AgiLex dual-arm manipulation tasks. As illustrated in ~\Cref{fig:app_3d_trajectory_cases_4} -~\Cref{fig:app_3d_trajectory_cases_7}, the model demonstrates robust performance in generating precise 3D spatial traces from complex natural language instructions. For example,
~\Cref{fig:app_3d_trajectory_cases_4} highlights the model's fine-grained spatial discrimination and reasoning abilities. In tasks such as \textit{Click Bell}, \textit{Click Alarmclock}, and \textit{Blocks Ranking}, RoboBrain 2.5 accurately identifies specific targets based on relative spatial descriptions (\eg, ``closest to milk box'', ``to the left of globe'') and comparative attributes (\eg, ``second largest'', ``farthest from the camera''). The generated traces correctly guide the agent to the intended objects among multiple distractors.
~\Cref{fig:app_3d_trajectory_cases_5} showcases the model's proficiency in dual-arm coordination and object manipulation. In scenarios like \textit{Handover Block}, \textit{Handover Mic}, \textit{Hanging Mug}, and \textit{Move Can Pot}, the model predicts coherent spatial traces that effectively handle arm-specific instructions (\eg, ``with the right arm'') and precise placement goals (\eg, ``hang it onto the rack''). These examples confirm that RoboBrain 2.5 can successfully ground high-level semantic instructions into collision-free, metric-aware trajectories for high-DoF robotic systems.

\vspace{0.3em}

These examples collectively demonstrate that RoboBrain 2.5’s precise 3D spatial reasoning capability effectively bridges the gap between natural language instructions and physical execution, enabling precise, constraint-compliant manipulation in real-world settings.

\vspace{-2.5em}

\vspace{-2.5em}

\vspace{-2.5em}

\vspace{-2.5em}

\vspace{-2.5em}

\vspace{-2.5em}

\vspace{-2.5em}

\clearpage

\subsection{Examples on Temporal Value Estimation}

\label{app:vis}
In this section, we provide additional qualitative visualizations to further substantiate the effectiveness and robustness of our method. We focus on three key aspects: the generalization of RoboBrain 2.5 across diverse semantic tasks, the temporal robustness of progress estimation under varying sampling intervals, and the trajectory visualization of real-world RL.

\vspace{0.3em}

\begin{figure*}[t]
    \centering
    \includegraphics[width=0.90\linewidth]{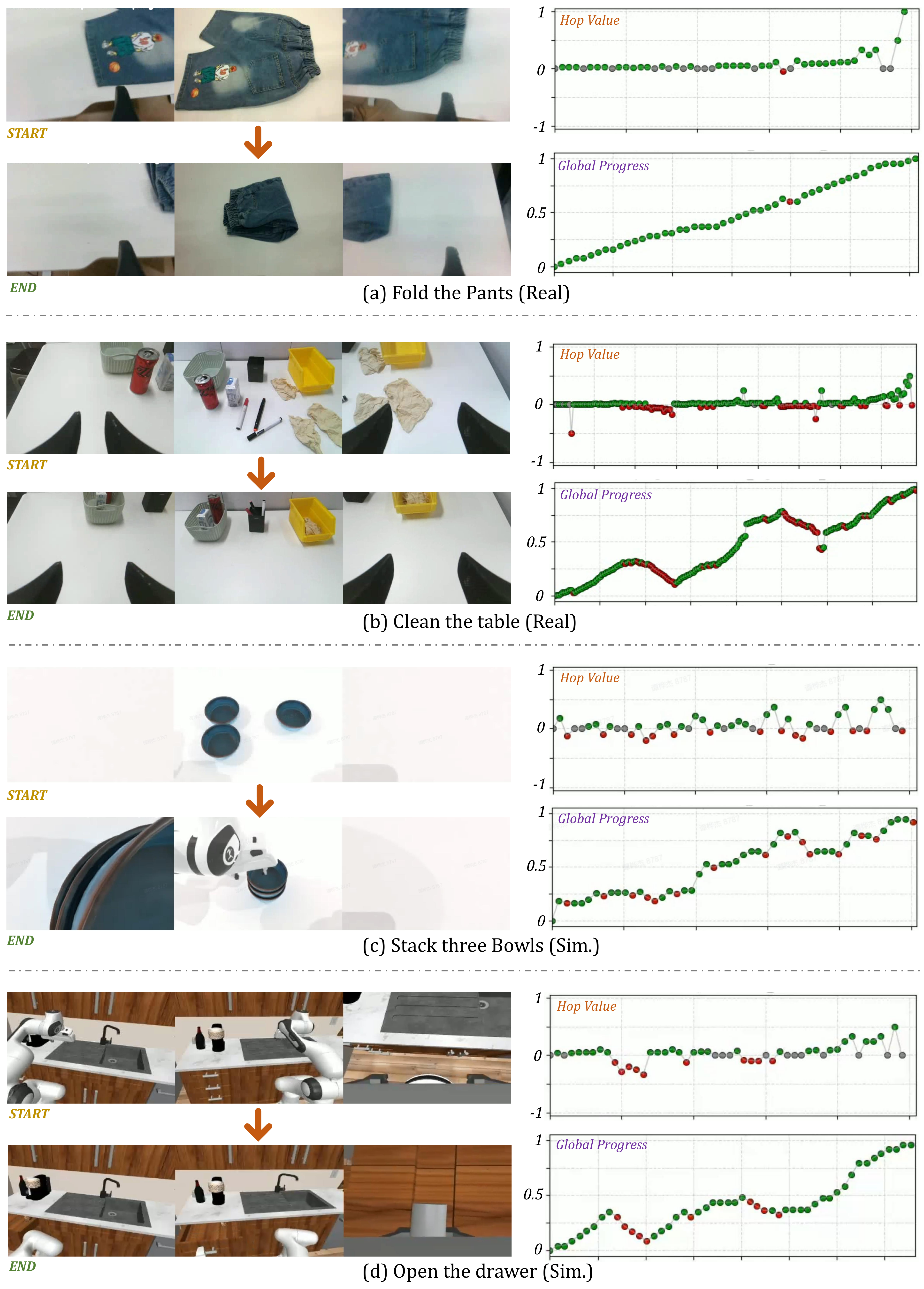}
    \caption{\textbf{RoboBrain 2.5 Progress Predictions across Diverse Tasks.} We visualize the frame-wise \textit{Hop} (instantaneous change) and accumulated \textit{Progress} predicted by RoboBrain 2.5 on unseen validation tasks.}
    \label{fig:vis_tasks}
\end{figure*}

\textbf{Dense Value Predictions on Diverse Tasks.}
~\Cref{fig:vis_tasks} illustrates the predicted Hop and Progress curves generated by our RoboBrain 2.5 across a wide spectrum of manipulation tasks, encompassing both real-world scenarios and simulation environments. Specifically, we visualize complex tasks such as deformable object manipulation (e.g., \textit{Fold the Pants}), unstructured real-world interaction (e.g., \textit{Clean the table}), and precise multi-stage simulation tasks (e.g., \textit{Stack three Bowls}, \textit{Open the drawer}). The results demonstrate that our model effectively generalizes across these distinct domains. In successful trajectories, the model consistently predicts positive Hop values corresponding to effective state transitions, which accumulate into a smooth, monotonic Progress curve that accurately reflects task completion. Crucially, the model is able to distinguish between effective progress and background noise, providing a stable signal even in visually cluttered real-world settings.

\vspace{0.3em}

\begin{figure*}[t]
    \centering
    \includegraphics[width=0.98\linewidth]{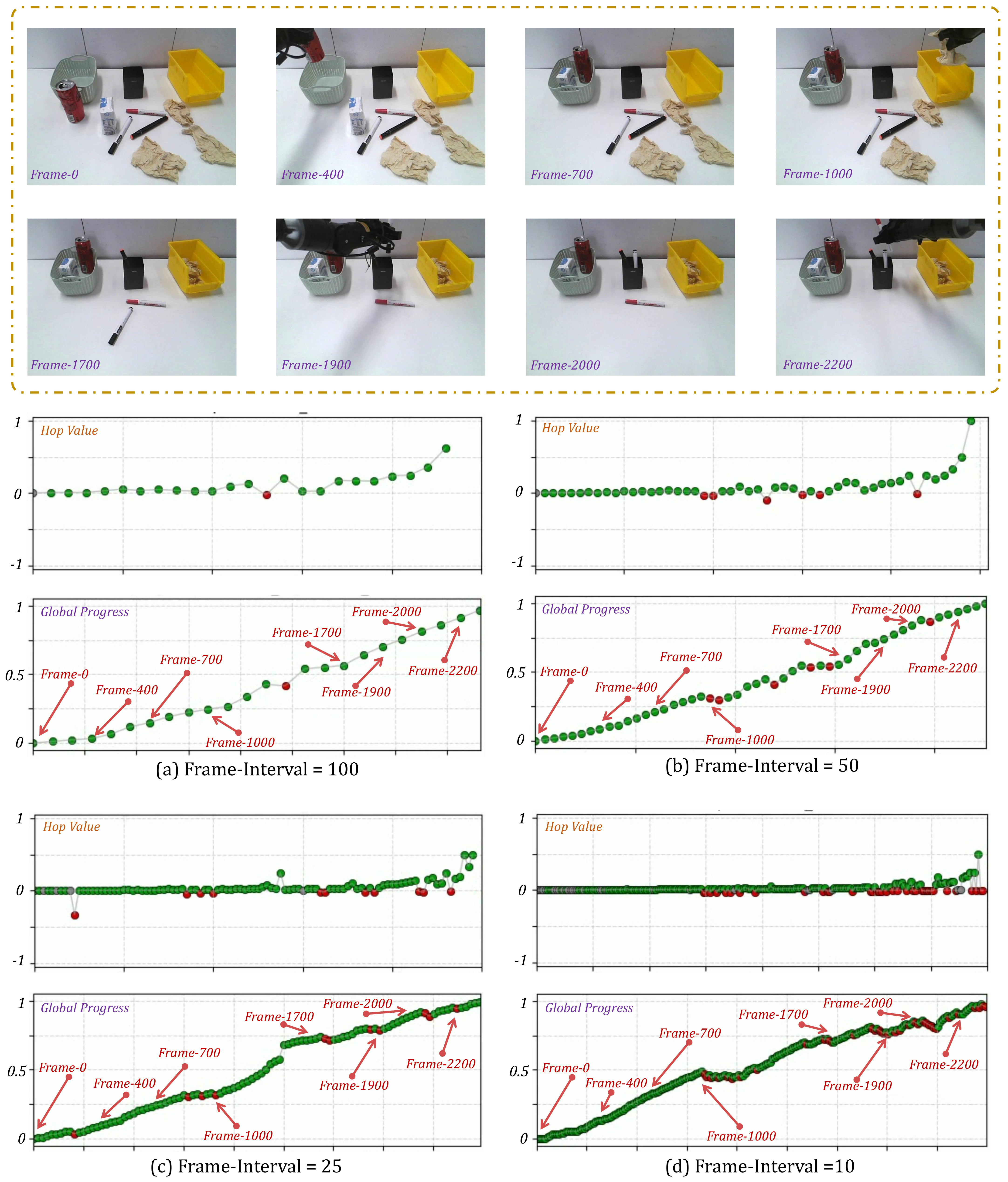}
    \caption{\textbf{Progress Estimation Consistency across Sampling Intervals.} We plot the reconstructed progress curves for the same trajectory using different frame strides (10, 25, 50, and 100 frames). The high overlap between curves demonstrates that our RoboBrain 2.5 is robust to temporal granularity and does not simply overfit to a specific frame rate.}
    \label{fig:vis_intervals}
\end{figure*}

\textbf{Robustness to Temporal Intervals.}
A robust reward model should remain consistent regardless of the video sampling rate or the control frequency of the robot. ~\Cref{fig:vis_intervals} provides a comparative analysis of RoboBrain 2.5's progress estimation when inputs are sampled at significantly different intervals ($\Delta t = 10, 25, 50, 100$ frames). As shown in the comparison, although the visual disparity between adjacent frames increases drastically with larger $\Delta t$, the model adaptively predicts larger Hop values to account for the increased semantic distance. Consequently, the reconstructed global progress curves across all sampling rates exhibit high alignment and overlap. This invariance highlights the model's ability to decouple physical progress from temporal duration, ensuring that the value estimation remains reliable whether the agent operates at high frequency or processes sparse keyframes.

\vspace{0.3em}

\begin{figure*}[t]
    \centering
    \includegraphics[width=0.80\linewidth]{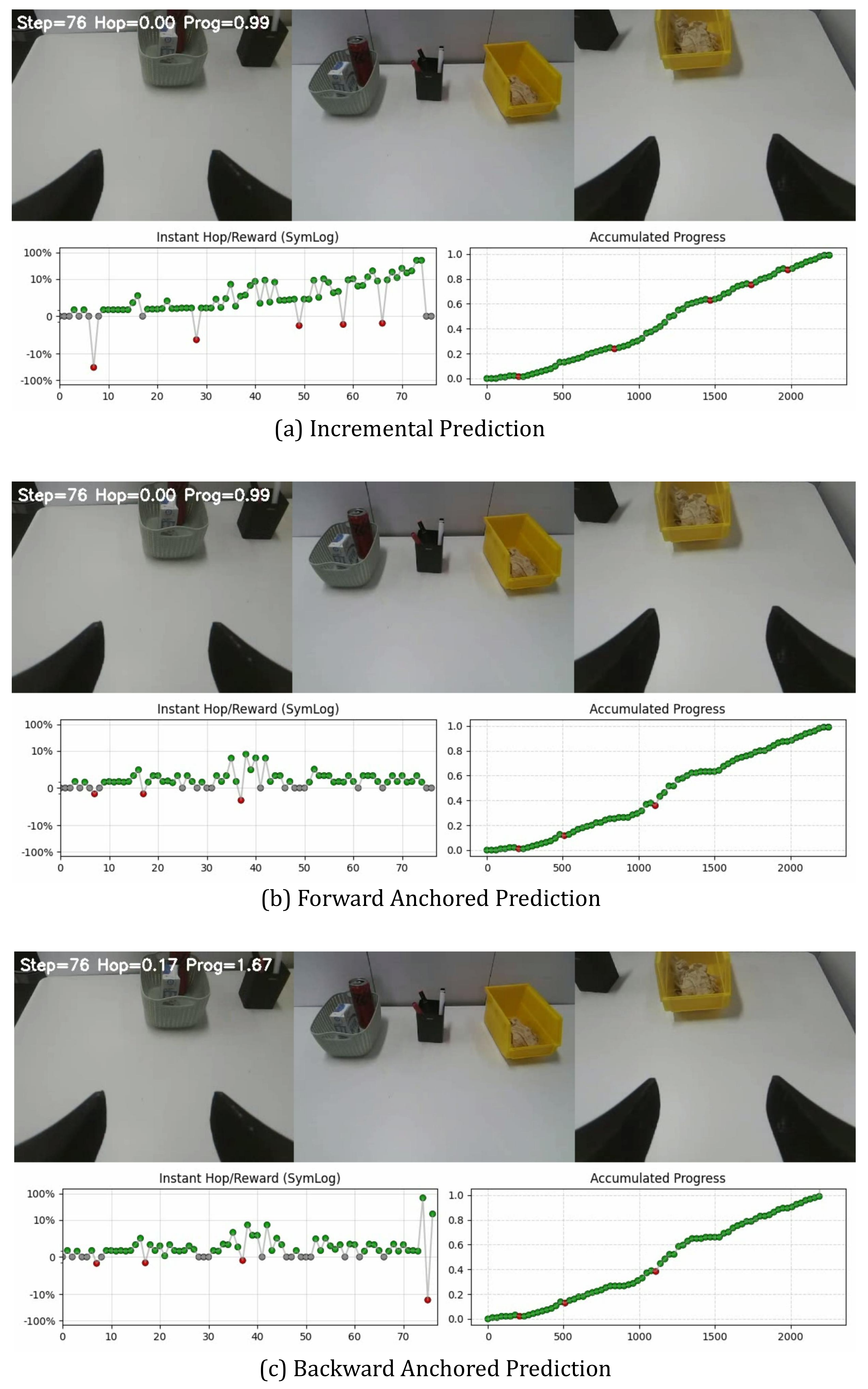}
    \caption{\textbf{RoboBrain 2.5 Progress Predictions across three modes.} We visualize the frame-wise \textit{Hop} (instantaneous change) and accumulated \textit{Progress} predicted by RoboBrain 2.5 with incremental, forward-anchored and backward-anchored mode.}
    \label{fig:progress_modes}
\end{figure*}

\textbf{Visualization of Different Progress Estimation Modes.}
Furthermore, to elucidate the robustness of our progress estimation, ~\Cref{fig:progress_modes} visualizes the three complementary perspectives used in our Multi-Perspective Progress Fusion strategy. Specifically, (a) \textit{Incremental Prediction} recursively accumulates frame-wise hop values to capture fine-grained local dynamics; (b) \textit{Forward-Anchored Prediction} estimates progress relative to the initial state, providing a stable baseline during early execution; and (c) \textit{Backward-Anchored Prediction} measures progress against the goal state, offering high sensitivity near task completion. As demonstrated, all three modes yield consistent, monotonic progress curves on unseen validation tasks, confirming that fusing these perspectives effectively mitigates error accumulation while maintaining global consistency.

\vspace{0.3em}

\begin{figure*}[t]
    \centering
    \includegraphics[width=0.98\linewidth]{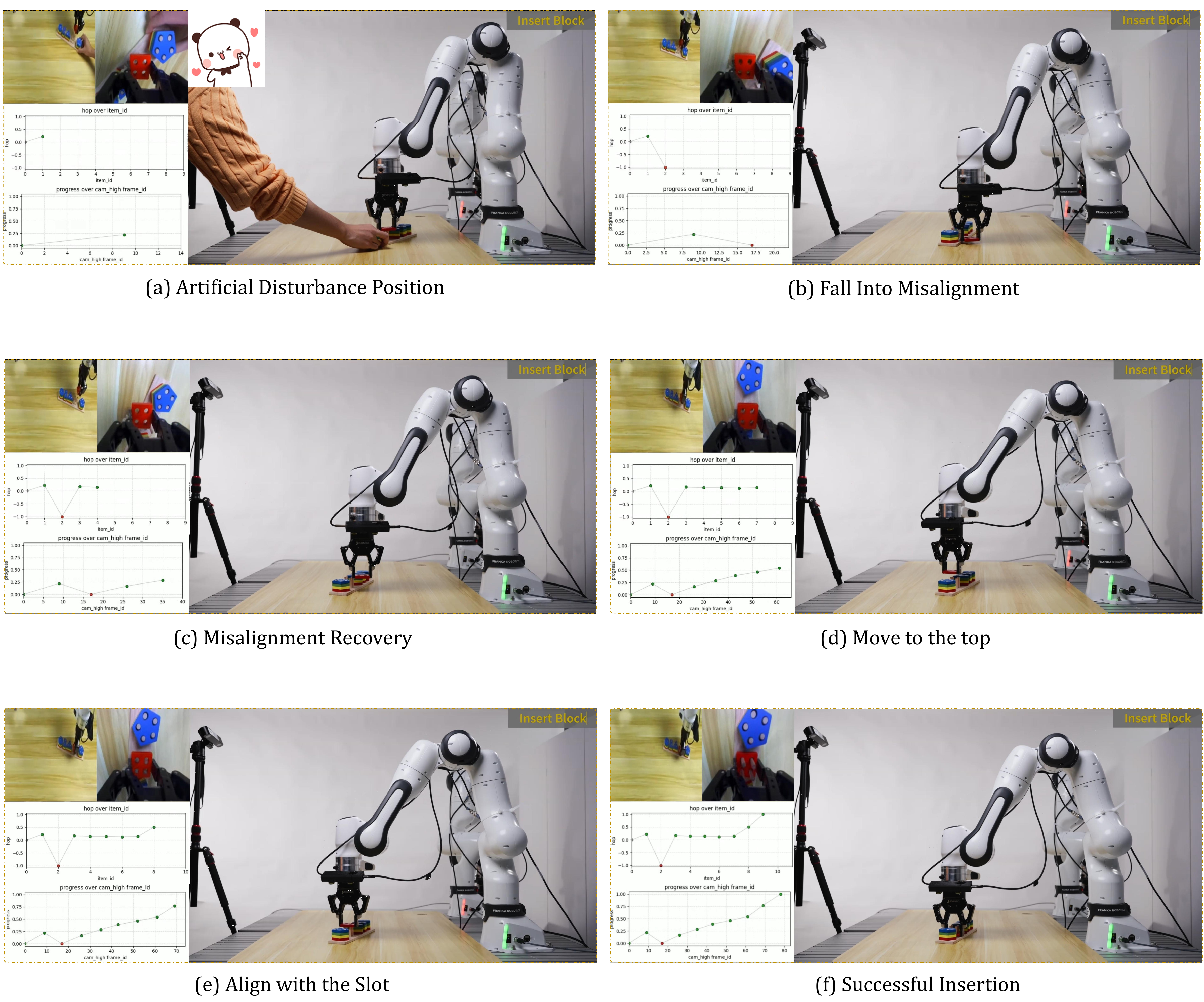}
    \caption{\textbf{Robustness to Artificial Disturbance during Real-World Execution.} We visualize a rollout of the converged policy (success rate $>95\%$) under human interference. Each sub-figure shows the third-person view, the ego-centric view, and the real-time RoboBrain 2.5 inference (Top: \textit{Hop}, Bottom: \textit{Progress}).
    \textit{\textbf{(a) Artificial Disturbance Position:}} A human hand intervenes and shifts the target board while the robot attempts to approach.
    \textit{\textbf{(b) Fall Into Misalignment:}} The robot misses the new position. Note that the RoboBrain 2.5 \textit{Progress} curve drops significantly (indicated by the red dot in the bottom inset), reflecting the failure state.
    \textit{\textbf{(c) Misalignment Recovery:}} The policy reacts to the visual feedback and the drop in reward, adjusting the end-effector position.
    \textit{\textbf{(d) Move to the top:}} The robot realigns directly above the target slot.
    \textit{\textbf{(e) Align with the Slot:}} Precise fine-tuning before insertion.
    \textit{\textbf{(f) Successful Insertion:}} The task is completed, with the progress estimation reaching its peak.}
    \label{fig:vis_rl}
\end{figure*}

\textbf{Real-World RL Rollout Visualization.}
Finally, ~\Cref{fig:vis_rl} visualizes the robustness of the policy learned for the ``Insert Block'' task. The policy used in this rollout was trained for approximately 20 minutes and achieved a success rate of over 95\%. To evaluate the policy's reactivity and the reward model's accuracy, we introduced an artificial disturbance during execution. As shown in the sequence, a human operator manually moves the target slot while the robot is in motion (a). This intervention causes the robot to miss the target and fall into misalignment (b). Crucially, the inset plots show that RoboBrain 2.5 immediately reflects this setback: the estimated \textit{Progress} drops sharply, correctly identifying that the state has regressed from the goal. This accurate negative feedback guides the agent to adjust its trajectory. The robot successfully recovers from the misalignment (c), repositions itself above the target (d), aligns with the slot (e), and completes the insertion (f). This demonstrates that RoboBrain 2.5 provides dense, semantically meaningful rewards that enable the agent to recover from unexpected external perturbations.

\clearpage

\section{Proof of Bounded Global Progress}
\label{app:proof}

In this subsection, we provide a formal proof that iteratively applying the predicted relative progress hops guarantees that the reconstructed global progress $\Phi^{\star}(s)$ remains strictly within the bounds $[0, 1]$, provided that the initial state is bounded and the model predictions lie within $[-1, 1]$.

\vspace{0.3em}

First, we define the general recursive update rule. Based on the definition of the hop label $\mathcal{H}(s_p, s_q)$ in Equation~\ref{eq:hop_calculation}, we derive the recursive update rule for estimating the global progress of the next state $\Phi^{\star}(s_t)$ given the current state $\Phi^{\star}(s_{t-1})$ and the predicted hop $H = \mathcal{H}(s_{t-1}, s_t)$. We assume the normalization where $\Phi(s_0)=0$ and $\Phi(s_M)=1$. Rearranging the equation, the update rule is:
\begin{equation}
\label{eq:update_rule}
\Phi^{\star}(s_t) = 
\begin{cases} 
    \Phi^{\star}(s_{t-1}) + H \cdot [1 - \Phi^{\star}(s_{t-1})] & \text{if } H \ge 0 \\
    \Phi^{\star}(s_{t-1}) + H \cdot \Phi^{\star}(s_{t-1}) & \text{if } H < 0
\end{cases}
\end{equation}

\vspace{0.3em}

Given that the initial progress $\Phi^{\star}(s_0) = 0$ and the predicted hop $H \in [-1, 1]$, the reconstructed global progress $\Phi^{\star}(s_t)$ satisfies $\Phi^{\star}(s_t) \in [0, 1]$ for all steps $t$.

\vspace{0.3em}

We proceed by mathematical induction as follow:
\textit{(1) Base Case ($t=0$):}
By definition, $\Phi^{\star}(s_0) = 0$, which satisfies $0 \in [0, 1]$.
\textit{(2) Inductive Step:}
Assume that for step $t-1$, the hypothesis holds: $0 \le \Phi^{\star}(s_{t-1}) \le 1$.
Let $G = \Phi^{\star}(s_{t-1})$ for brevity, where $G \in [0, 1]$. We analyze the next state $\Phi^{\star}(s_t)$ under two cases (\textit{i.e.,} Positive Hop and Negative Hop) based on the sign of the predicted hop $H$.

\vspace{0.3em}

\begin{itemize}
    \item \textbf{\textit{Case 1: Positive Hop (Progress), $0 \le H \le 1$.}} \\
    From~\Cref{eq:update_rule}, the update is written as:
    \begin{equation}
        \Phi^{\star}(s_t) = G + H(1 - G)
    \end{equation}
    Rearranging terms to view this as a convex combination:
    \begin{equation}
        \Phi^{\star}(s_t) = H + G(1 - H)
    \end{equation}
    
    \textit{Lower Bound:} Since $G \ge 0$, $H \ge 0$, and $(1-H) \ge 0$, it follows that $\Phi^{\star}(s_t) \ge 0$.
    
    \textit{Upper Bound:} Since $G \le 1$, we substitute the maximum value of $G$:
    \begin{align*}
        \Phi^{\star}(s_t) &= H + G(1 - H) \\
        &\le H + 1 \cdot (1 - H) \\
        &= H + 1 - H \\
        &= 1
    \end{align*}
    Thus, $0 \le \Phi^{\star}(s_t) \le 1$ when $H \ge 0$.

    \item \textit{\textbf{Case 2: Negative Hop (Regress), $-1 \le H < 0$.}} \\
    From~\Cref{eq:update_rule}, the update is:
    \begin{equation}
        \Phi^{\star}(s_t) = G + H \cdot G = G(1 + H)
    \end{equation}
    
    \textit{Lower Bound:} Since $H \in [-1, 0)$, the term $(1 + H) \ge 0$. Since $G \ge 0$, the product $G(1+H) \ge 0$.
    
    \textit{Upper Bound:} Since $H < 0$, the term $(1 + H) < 1$. Combining this with $G \le 1$:
    \begin{equation*}
        \Phi^{\star}(s_t) = G(1 + H) \le 1 \cdot (1) = 1
    \end{equation*}
    Thus, $0 \le \Phi^{\star}(s_t) \le 1$ when $H < 0$.
\end{itemize}

\vspace{0.3em}

\noindent\textbf{Conclusion.} Since the property holds for the base case and is preserved in both update scenarios during the inductive step, we conclude that $\Phi^{\star}(s_t) \in [0, 1]$ for all $t$.

\end{document}